\documentclass{article}

\usepackage{microtype}
\usepackage{graphicx}
\usepackage{subfigure}
\usepackage{booktabs} 
\usepackage{float}
\usepackage{hyperref}

\usepackage[accepted]{icml2025}

\usepackage{amsmath}
\usepackage{amssymb}
\usepackage{mathtools}
\usepackage{amsthm}
\usepackage{tikz}
\usepackage{xspace}

\usepackage[capitalize,noabbrev]{cleveref}

\theoremstyle{plain}

\theoremstyle{definition}

\theoremstyle{remark}

\usepackage{enumitem}
\setlist[itemize]{leftmargin=*}

\usepackage{multirow}
\usepackage{graphicx}
\usepackage{pifont}
\usepackage{xcolor,colortbl}
\usepackage{multicol}
\usepackage{tablefootnote}
\usepackage[para,online,flushleft]{threeparttable}
\usepackage{xspace}

\usepackage{setspace}

\usepackage[many]{tcolorbox}

\usepackage{letltxmacro}

\usepackage{siunitx}
\usepackage{xcolor}
\usepackage{listings}

\def\rr{\mathbb{R}}

\definecolor{commentcolor}{RGB}{110,154,155} 
\usepackage[textsize=tiny]{todonotes}

\newcommand{\sysname}{\textsc{HeadInfer}\xspace}

\icmltitlerunning{\sysname: Memory-Efficient LLM Inference by Head-wise Offloading}

\begin{document}

\twocolumn[
\icmltitle{\sysname: Memory-Efficient LLM Inference by Head-wise Offloading}




\begin{icmlauthorlist}
\icmlauthor{Cheng Luo}{caltech}
\icmlauthor{Zefan Cai}{UW}
\icmlauthor{Hanshi Sun}{cmu}
\icmlauthor{Jinqi Xiao}{Rutgers}
\icmlauthor{Bo Yuan}{Rutgers}
\icmlauthor{Wen Xiao}{MS}
\icmlauthor{Junjie Hu}{UW}
\icmlauthor{Jiawei Zhao}{caltech}
\icmlauthor{Beidi Chen}{cmu}
\icmlauthor{Anima Anandkumar}{caltech}
\end{icmlauthorlist}

\icmlaffiliation{UW}{University of Wisconsin-Madison}
\icmlaffiliation{MS}{Microsoft}
\icmlaffiliation{Rutgers}{Rutgers University}
\icmlaffiliation{caltech}{California Institute of Technology}
\icmlaffiliation{cmu}{Carnegie Mellon University}

\icmlcorrespondingauthor{Cheng Luo}{chengluo@caltech.edu}
\icmlcorrespondingauthor{Anima Anandkumar}{anima@caltech.edu}


\icmlkeywords{Machine Learning, ICML}

\vskip 0.3in
]



\printAffiliationsAndNotice{}  

\begin{abstract}
Transformer-based large language models (LLMs) demonstrate impressive performance in long context generation. Extending the context length has disproportionately shifted the memory footprint of LLMs during inference to the key-value cache (KV cache). In this paper, we propose \sysname, which offloads the KV cache to CPU RAM while avoiding the need to fully store the KV cache for any transformer layer on the GPU. \sysname employs a fine-grained, head-wise offloading strategy, maintaining only selective attention heads' KV cache on the GPU while computing attention output dynamically. Through roofline analysis, we demonstrate that \sysname maintains computational efficiency while significantly reducing memory footprint. We evaluate \sysname on the Llama-3-8B model with a 1-million-token sequence, reducing the GPU memory footprint of the KV cache from 128 GB to 1 GB and the total GPU memory usage from 207 GB to 17 GB, achieving a 92\% reduction compared to BF16 baseline inference. Notably, \sysname enables 4-million-token inference with an 8B model on a single consumer GPU with 24GB memory (e.g., NVIDIA RTX 4090) without approximation methods.
\end{abstract}

\vspace{-2em}
\section{Introduction}
\label{submission}

Modern Large Language Models (LLMs) increasingly support extremely long inputs: Llama-3 \cite{dubey2024llama} handles up to 128K tokens, Claude \cite{anthropic2024claude} supports up to 1 million tokens, while Gradient AI \cite{gradientlongcontextllama3} extends Llama-3 to 4 million tokens. These extended context lengths improve performance on tasks such as book summarization \cite{pal2023giraffe} and video generation \cite{liu2024world}, requiring millions of tokens.

\begin{figure}[t]
\begin{center}
  \includegraphics[scale=0.26]
  {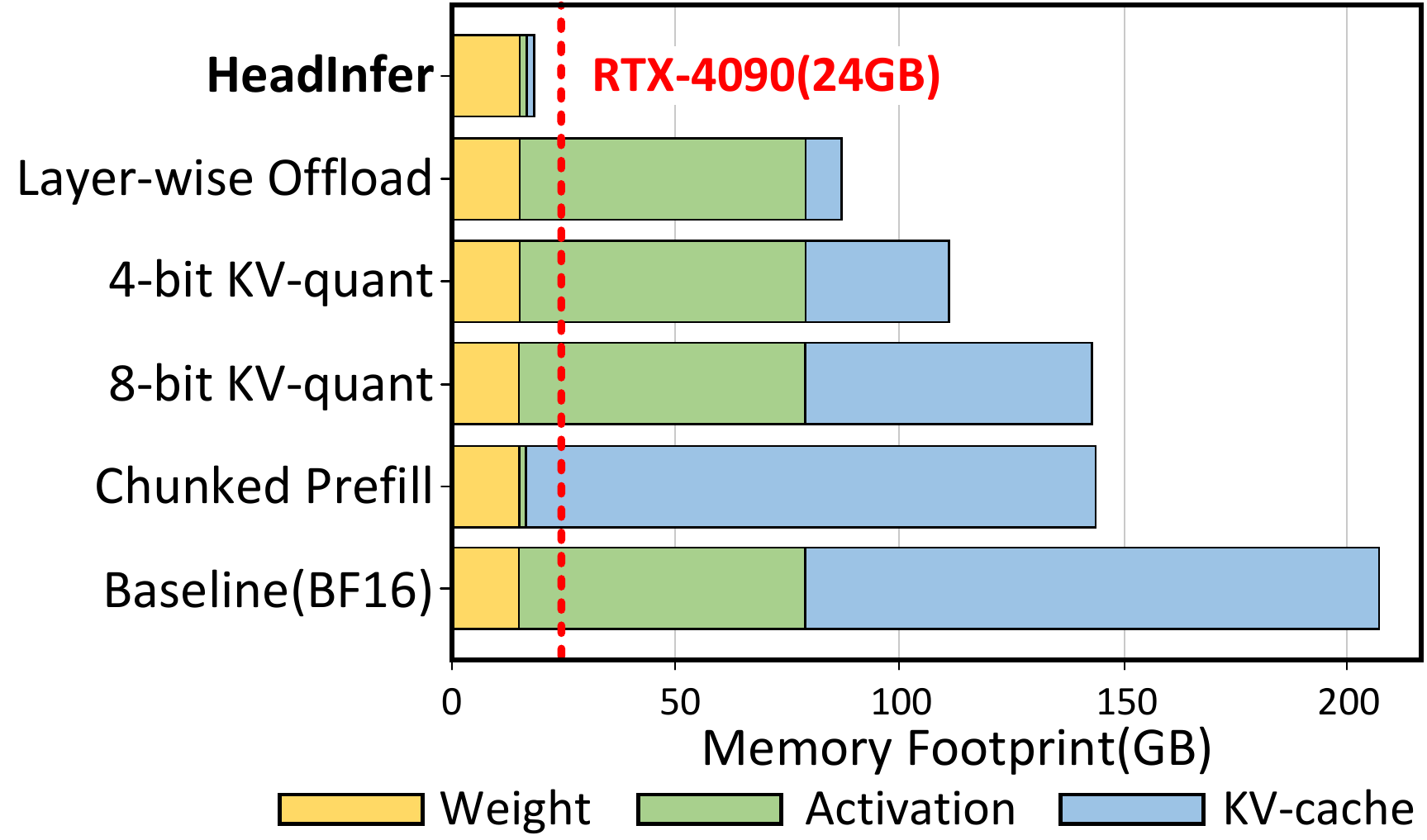}
  \vspace{-1.25em}
  \caption{Estimated memory consumption of inference a Llama-3-8B model with 1 million token on a single GPU.}
  \label{fig:first_analysis}
  \vspace{-2.5em}
\end{center}
\end{figure}

As context lengths increase, memory usage and latency grow significantly due to self-attention in transformers. To improve inference efficiency \cite{zhou2024survey}, most LLM inference consists of two phases: prefill for input processing and decoding for token generation, with key and value states from attention cached for reuse (KV cache). However, as the context length increases, the KV cache memory grows rapidly, posing significant challenges for storage and efficiency. For example, generating 1 million tokens using a Llama-3-8B model requires 207 GB of memory (15 GB for pre-trained parameters, 64 GB for activation, and 128 GB for KV cache \protect\footnotemark[1]) as shown in Figure \ref{fig:first_analysis}. This makes long-context inference unfeasible on consumer-level GPUs such as the NVIDIA RTX-4090 with 24 GB of memory.

\footnotetext[1]{The calculation is based on standard inference generation
using Hugging Face with BF16 precision and a batch size of 1.
Details of how it is calculated can be referred to in Appendix \ref{Memory Analysis of Inference LLama3-8B}.}

\begin{figure*}[t]
\begin{center}
  \includegraphics[scale=0.5]
  {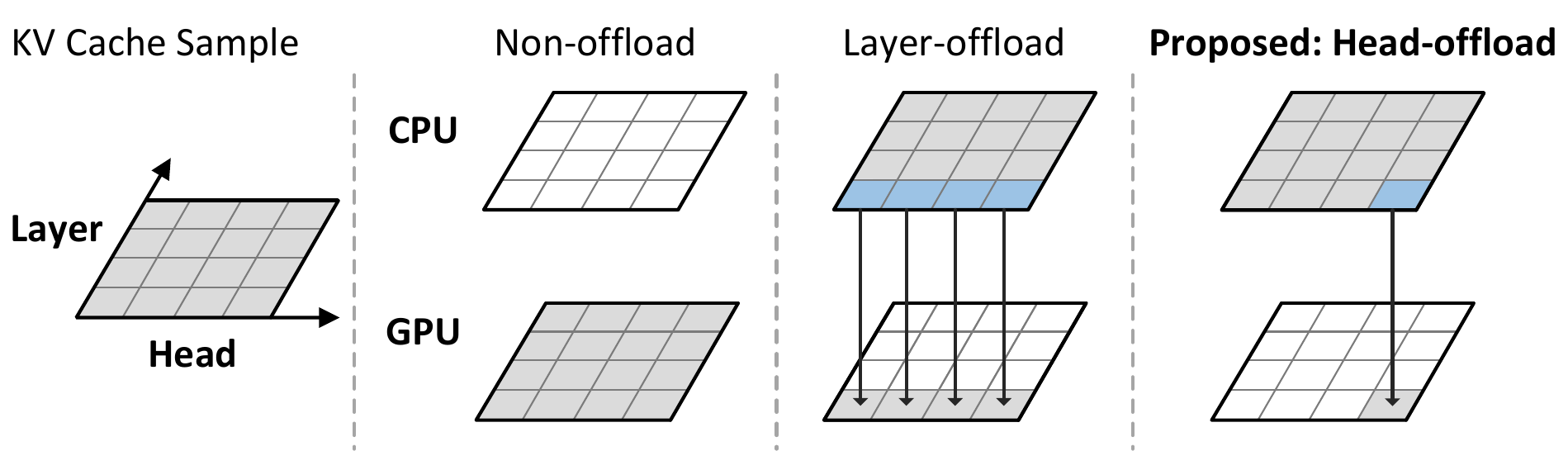}
  \vspace{-0.5em}
  \caption{Demonstrations of KV cache policies in inference. Full KV cache contains two main dimensions: layer and head. Layer-wise offloads KV cache in the layer's dimension, with a cache budget of all heads per layer. \sysname further reduces GPU memory by adaptively reallocating cache budgets in the head's dimension, with a cache budget of one head.}
  \label{fig:diagram}
  \vspace{-1.5em}
\end{center}
\end{figure*}

To address this memory challenge, modern LLM serving systems have introduced offloading strategies that transfer data to CPU memory, enabling efficient LLM inference within hardware constraints \cite{sheng2023flexgen, lee2024infinigen, aminabadi2022deepspeed}. By moving portions of the KV cache to the CPU, these methods allow for much longer context generation than would otherwise fit in GPU memory. However, these methods can hardly be applied to consumer-level GPUs such as the RTX-4090, where only 8GB of memory remain for the KV cache and activations. This is due to two factors: (1) offloading does not reduce activation memory usage, and (2) because offloading is applied at a coarse granularity, the remaining KV cache on the GPU ends up larger than the rest of the memory usage.

\textbf{Our approach}: We propose head-wise offload (\sysname), an inference framework that drastically reduces the GPU memory needed for LLM inference by offloading the KV cache at the level of individual attention heads. The key idea is to leverage attention head independence to decompose the attention computation in a head-wise manner, requiring only one head of the KV cache on the GPU at a time and offloading the rest to the CPU. As shown in Figure \ref{fig:diagram}, \sysname frees the GPU system from storing the KV cache of the entire model or layer and instead stores only a single head at any given time. This fine granularity slashes the GPU memory footprint while maintaining exact mathematical equivalence.

To enable efficient long-context inference on consumer GPUs, \sysname integrates several key optimization techniques: adaptive heads grouping, chunked prefill, and asynchronous data transfer. Adaptive head grouping dynamically adjusts the number of attention heads retained on the GPU, progressively reducing the memory footprint as the context length increases. Chunked prefill reduces peak memory usage by splitting long input sequences into smaller segments, processing them incrementally to avoid excessive activation memory consumption. Asynchronous data transfer overlaps KV cache movement between GPU and CPU and ensures computation proceeds without stalls from memory transfers. 


Through roofline analysis \cite{williams2009roofline}, we demonstrate that \sysname maintains high compute efficiency while significantly reducing GPU memory consumption. 
We implement \sysname on the Hugging Face framework and evaluate it on representative LLMs, including Llama, Qwen, Mistral, and Gemma, with varying model sizes and sequence lengths. More than the 1 million token inferences shown in Figure \ref{fig:first_analysis}, our results demonstrate that \sysname extends the Llama-3-8B model's context length from 25K (standard inference) and 45K (standard offload) to 4 million tokens, achieving around 100x improvement in context length extension using NVIDIA RTX 4090.

In summary, this paper presents the following contributions.

\begin{itemize}[leftmargin=*]
\item \sysname enables inference with context length with over a million tokens on a single consumer GPU like NVIDIA RTX 4090.

\item In-depth analysis using the roofline model: high compute efficiency is achieved during prefill without shifting computation into the memory-bound regime.

\item Fully general and implementation-agnostic attention: \sysname supports
dense as well as sparse attention, and it works with head-wise sparse attention such as duo-attention \cite{xiao2024duoattention}.

\item Support for massive model inference: \sysname collaborates with pipeline parallelism to support larger models like Llama3-70B with 1 million tokens.

\item Easy-to-use and portable, requiring minimal code changes to the existing inference frameworks. The code example can be referred to in Appendix \ref{sec:implementation}.




\end{itemize}

\section{Related Work}

\subsection{Generative Inference and KV Caching}

Generative LLM inference typically involves two main stages: the  \emph{prefill} stage and the  \emph{decoding} stage. In the prefill stage, the model processes the initial input prompt by computing attention scores for all tokens in the input sequence. 
For long-context input, it is common to adopt \emph{chunked prefill} \cite{agrawal2024taming}, which divides the prompt into fixed-length chunks to incrementally build the KV cache. This strategy significantly reduces peak memory usage by lowering linear layers' peak intermediate activation size from the entire sequence to just the smaller chunk. 
In the subsequent decoding stage, each newly generated token from the prefill stage is fed back into the model, creating an autoregressive generation process. The LLM produces one new token at a time, and each token attends to all previous KV cache. 

\subsection{Lossy KV Cache Management} Evit KV cache can reduce memory usage and computational complexity. One direction is to identify and retain only the most 'valuable' tokens within the KV cache. Representative methods include Sliding Window Attention \cite{beltagy2020longformer}, Heavy Hitter \cite{zhang2024h2o}, and StreamingLLM \cite{xiao2023efficient}. Another direction is to identify and retain the attention heads. Wu et al. \cite{wu2024retrieval} find a way to evaluate the importance of attention heads. Head-wise sparsity such as duo-attention \cite{xiao2024duoattention}, HeadKV \cite{fu2024not}, and Razorattention \cite{tang2024razorattention} start to divide up KV cache budgets based on the importance of each head, which is usually determined by the need for retrieval or reasoning.  Minference \cite{jiang2024minference} takes this idea further by applying distinct sparse patterns to different heads.

\subsection{Offloading KV Cache}

Offloading the KV cache from the GPU memory to CPU DRAM is another memory-efficient strategy. For instance, LayerKV \cite{xiong2024layerkv} implements efficient layer-wise KV offloading and overlapping data transfers to improve the context length. FastDecode \cite{he2024fastdecode} and NEO \cite{jiang2024neo} also offload parts of the KV cache to the CPU and perform attention computations on the CPU. ShadowKV \cite{sun2024shadowkv} combines SVD decomposition with offloading to reduce communication overhead. FlexInfer \cite{xu2024vtensor} introduces the vTensor abstraction to better manage heterogeneous memory resources. Infinigen \cite{lee2024infinigen} introduces dynamic KV cache management with offloading systems.

\section{\sysname: Head-wise Offload}
\subsection{Background:}
\label{sec:figs}


\textbf{Regular Inference with KV cache Generation.} At each time step $t$, the input token
$x_t$ is transformed into an embedding vector $E(x_t)$ by embedding the token. Then, a linear projection generates the key
$K_t$ and the value $V_t$, which can be written as follows:
\begin{equation}
    K_t = W_K  E(x_t), V_t = W_V E(x_t)
\end{equation}
Here $W_K$ and $V_t$ are projection weights, and the current $K_t$ and  $V_t$ are appended to the existing key cache $K_{cache}$ and value cache $V_{cache}$. 
\begin{equation}
    K_{cache} = [K_1, K_2, ..., K_t], V_{cache} = [V_1, V_2, ..., V_t]
\end{equation}

The state of the cache can be memory-intensive.
Together with sequence length $S$ and hidden length $D$, $K_t,V_t \in \rr^{S\times H} $. This takes $2 \times S \times D$ memory. 

\textbf{Attention with KV Cache.} During the computation of self-attention at time step $t$, the model utilizes the entire sequence of keys and values from time steps $1$ to $t$ (stored in the KV cache) alongside the new query $Q_t$:
\begin{equation}
    A_t = Softmax(\frac{Q_t K_{cache}^T}{\sqrt{d_k}}) V_{cache}
\end{equation}
where $d_k$ is the dimensionality of the keys.

\textbf{Offload KV Cache.}

When the context length $S$ grows on a million scale, or the GPU’s on-device High-Bandwidth Memory (HBM, or $M_{HBM}$) is small on the consumer GPU, it may become insufficient to store the entire key-value cache. In such scenarios, KV cache offloading provides a practical approach to handling memory limitations. Offloading can temporarily move parts of the KV cache to CPU RAM or other external resources (e.g., NVMe storage and CPU disk). However, each offloading strategy introduces potential communication overheads that must be carefully weighed against the gains in usable context length.

For a batch of size $B$, a transformer with $L$ layers, and the KV cache with the bytes per element $sizeof(datatype)$, the total KV cache size is:
\begin{equation}
    Size_{KV{cache}} = 2 \times B \times L \times S \times D \times sizeof(datatype)
\end{equation}

If $ Size_{KV_{cache}} > M_{HBM}$, the system can offload some portion of the KV cache to external memory to avoid out-of-memory errors. Let $\alpha (0 \leq \alpha \leq 1)$ be the fraction of the KV cache that remains on the GPU, and $1 - \alpha$ be the fraction offloaded to external memory.  The memory footprint on the GPU $Size_{on-GPU}$ can be expressed as follows:
\begin{equation}
    Size_{on-GPU} = \alpha \times Size_{KV{cache}}
\end{equation}
Therefore, we require:
\begin{equation}
    Size_{on-GPU} \leq M_{HBM}, \alpha \leq \frac{M_{HBM}}{Size_{KV_{cache}}}
\end{equation}

\begin{figure*}[t]
\begin{center}
  \includegraphics[scale=0.49]
  {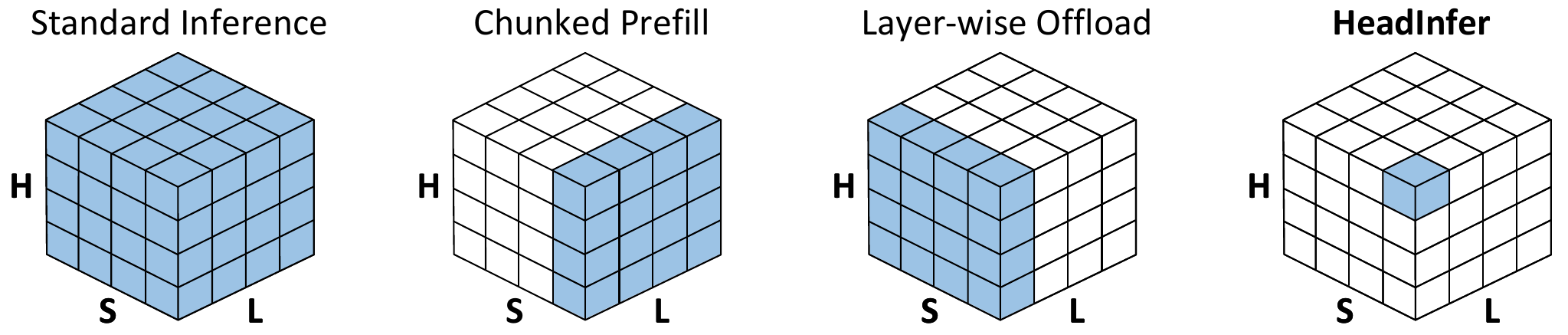}
  \caption{Granularity of different methods. Each cube represents the entire attention process along three dimensions: Sequence (S), Layers (L), and Heads (H). Standard inference puts everything on the GPU.
Chunked-prefill fetches only a part of the sequence dimension of all tokens on the GPU at a time.
Layer-wise offloading fetches a subset of layers on the GPU, offloading the rest.
\sysname introduces an even finer approach that maintains only selected heads within a layer. }
  \label{fig:Granularity}
  \vspace{-1.5em}
\end{center}
\end{figure*}

\subsection{Head-wise Offload (\sysname)}

\textbf{Head-wise KV Cache Generation.} In transformer architectures, each self-attention layer is split into multiple attention heads $H$. Each head has its own set of key-value (KV) tensors:
\begin{equation}
    K^{(h)} \in  \mathbb{R}^{S \times D_h}, V^{(h)} \in  \mathbb{R}^{S \times D_h}
\end{equation}

where $D$ is divided by $H$ so that each head handles a subdimension $D_h =D/H$.  Therefore, instead of a single large KV cache, a head-wise approach organizes cache in $H$ separate memory spaces. Formally, at time step $t$, we have:
\begin{equation}
    K_{cache}^{(h)} = [K_1^{(h)}, ..., K_t^{(h)}], V_{cache}^{(h)} = [V_1^{(h)}, ..., V_t^{(h)}]
\end{equation}

As a result, each head stores its keys and values in a contiguous memory space, enabling selective offloading of certain heads' cache when memory constraints emerge. 

During self-attention at the time step $t$, we calculate the attention output for each head $h$ independently, using:
\begin{equation}
    A_t^{(h)} = Softmax(\frac{Q_t^{(h)} K_{cache^{(h)}}^T}{\sqrt{d_k}}) V_{cache^{(h)}}
\end{equation}
Finally, the outputs ${A_t^{(h)}}^H_{h=1}$ are concatenated to form the final output of attention for that layer.

\textbf{Head-wise Offload.}
Since the attention computation has a head-wise independence, if we can keep the KV cache of a single head rather than the entire layer, then the memory consumption can be reduced substantially. This leads to
our proposed head-wise offload (\sysname) strategy.


\sysname is designed to minimize the fraction of on-GPU data $\alpha$ (the fraction of the total KV cache stored on GPU). Using Llama-3-8b as an example, we define $H_{all}$ as the total number of attention heads of a given model, which equals the number of heads per layer times the number of layers $H \times L$. $H_{on}$ is the number of heads $h$ retained on the GPU, and $H_{off}$ is the number of heads offloaded to external memory (CPU); obviously, we have
$H_{on} + H_{off} = H_{all}$

Define $\alpha$ as the fraction of the KV cache that remains on the GPU. We can store all KV cache on the GPU if $\alpha = 1$ or layer-wise offload if $\alpha = 1/L$. In our head-wise scheme:
\begin{equation}
    \alpha = \frac{H_{on}}{H_{all}} =\frac{1}{L\times H}
\end{equation}
Here we keep only a single head on the GPU, and the fraction of the total KV cache that occupies GPU memory is reduced by $\alpha = 1/(L\times H)$, with a total size of:
\begin{equation}
    S_{on-GPU} =  2 \times B  \times S \times  D_{H} \times sizeof(datatype)
\end{equation}
By reducing $\alpha$, we can preserve GPU memory capacity for extended context inference and make the million-level inferences on consumer GPUs possible.

\subsection{Granularity: Sequences, Layers, and Heads}

\sysname differs from traditional offload in terms of granularity. When deploying large language models, each dimension of the model can become a potential bottleneck in GPU memory. Naively offloading the entire KV cache or entire layers can be too coarse, leading to suboptimal memory usage.

As shown in Figure \ref{fig:Granularity}, HeadInfer addresses this challenge by introducing a hierarchical set of techniques, including \emph{chunked-prefill}, \emph{layer-wise offload}, and \emph{head-wise offload} that operate at increasingly fine-grained levels of sequence (S), layers (L), and heads (H). 


\begin{itemize}[leftmargin=*]
\item \textbf{Chunked-Prefill (S)}: Rather than processing all sequences of input tokens at once, \sysname divides the sequence into smaller chunks, each processed separately during the prefill stage. This partition helps reduce the activation memory usage.

\item \textbf{Layer-Wise Offload (L)}: Instead of storing the entire KV cache on the GPU, \sysname offloads it to the GPU and fetches the KV cache of specific layers on demand. Consequently, only the relevant portion of the KV cache resides on the GPU at any given time.

\item \textbf{Head-Wise Offload (H)}: Within each layer, \sysname can selectively offload the KV cache for all attention heads, fetching certain attention heads on demand. This offers the finest granularity by focusing on a subset of heads within each layer.
\end{itemize}

By combining these techniques, \sysname allows precise control over which parts of the activation and KV cache remain on the GPU. Because these dimensions nest naturally as chunks are subsets of the sequence, layers are subsets of the model, and heads are subsets of the attention layer; \sysname can flexibly adapt to a wide range of hardware constraints, such as available GPU memory, while meeting performance requirements.




\section{\sysname for Memory-Efficient Inference}

\begin{figure}[t]
\begin{center}
  \includegraphics[scale=0.365]
 {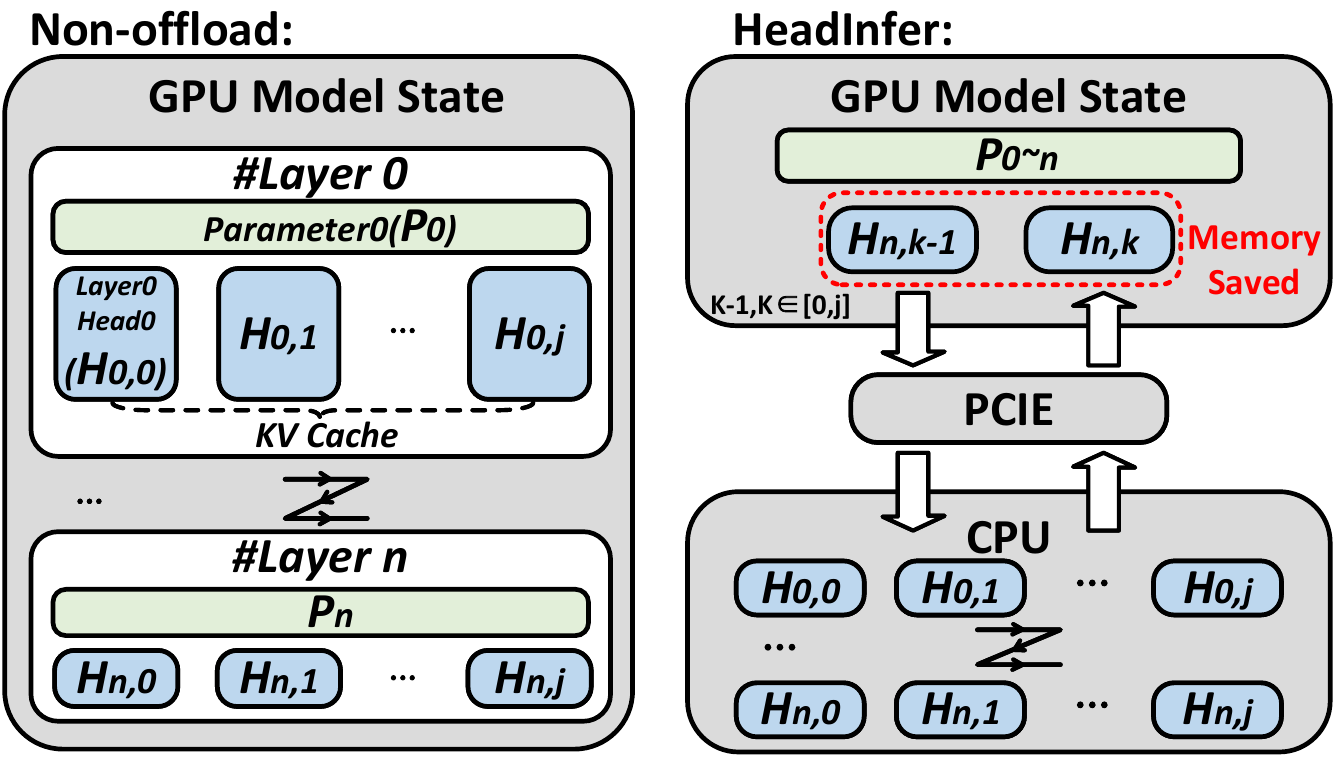}
 \vspace{-2em}
 \caption{\sysname snapshot. All parameters are stored on the GPU. Head-wise partitioned KV cache is moved across GPU and CPU with the ping-pong memory.}
 \label{fig:pipe}
 \vspace{-2em}
\end{center}
\end{figure}

For the offloading strategy, the potential communication overheads must be carefully considered against the gains in context length extension. This is because the offloading communication cost grows along with the context length. Fortunately, for this, \sysname can overlap the communication cost with evening growing attention computation. This is achieved by a ping-pong memory design (Figure \ref{fig:pipe}) for asynchronous offloading and prefetch. We present the implementation on algorithm \ref{alg:headinfer}.

\textbf{Overlapping Head-wise KV Cache Movement and GPU Compute.} The overlapping mechanism is critical for offload performance, especially for long-context inference when the computation time can completely overlap the offload communication. While the GPU computes attention for the current head $A_t^{(h)}$, it concurrently prefetches the next head's KV cache from the CPU and offloads the current head's cache back to the CPU. Ping-pong memory enables non-blocking PCIe transfers to synchronize memory movement with computation. This mechanism is presented in Algorithm \ref{alg:headinfer} lines 10-11 and 25-26, as async prefetch and async update. 

\textbf{Efficient Management of KV Cache.}
Effective KV cache management guarantees the feasibility of long context head-wise offload. Key strategies here include head-wise partitioning and pre-allocation. Head-wise partitioning makes sure that each head has its own KV cache $K^{(h)}$, $V^{(h)}$, and allows selective offloading and retention based on memory availability and head importance. \sysname pre-allocates the CPU's KV cache memory and the GPU's ping-pong memory to avoid runtime memory allocation overheads.

\textbf{Adaptive Head-wise Offloading.} Adaptive head-wise offloading reduces kernel launch overheads caused by head-wise partitioning, especially for small context sizes. This involves fusing multiple attention heads into a single kernel, reducing the number of kernel launches at the cost of larger KV cache fetches. For instance, we fuse all heads when processing the LLAMA3 inference 1-500K context, a process that is identical to the standard attention calculation. We divide the attention heads into two groups when processing 500K-1M, four groups when processing 1M-2M, and eight groups when processing 2M-4M. Specifically for eight groups, \sysname performs head-wise offload with finest granularity, where only one head is left on the GPU. 

\textbf{Extension: Combining with Head-wise Sparsity.} \sysname is compatible with existing sparse optimization techniques. Our work focuses on integrating \sysname with head-wise sparsity, such as duo-attention \cite{xiao2024duoattention}. This head-wise sparsity reduces memory by truncating less important heads to a fixed length (typically under 1k). We designate these heads as on-GPU $H_\text{on}$ heads without offloading since they consume minimal GPU memory, and offloading KV cache smaller than 1k lead to performance degradation, as analyzed in section \ref{Analysis}. The detailed design for the extension can be found in Appendix \ref{appx-implementation}

\begin{figure*}[t]
\begin{center}
  \includegraphics[scale=0.66]
  {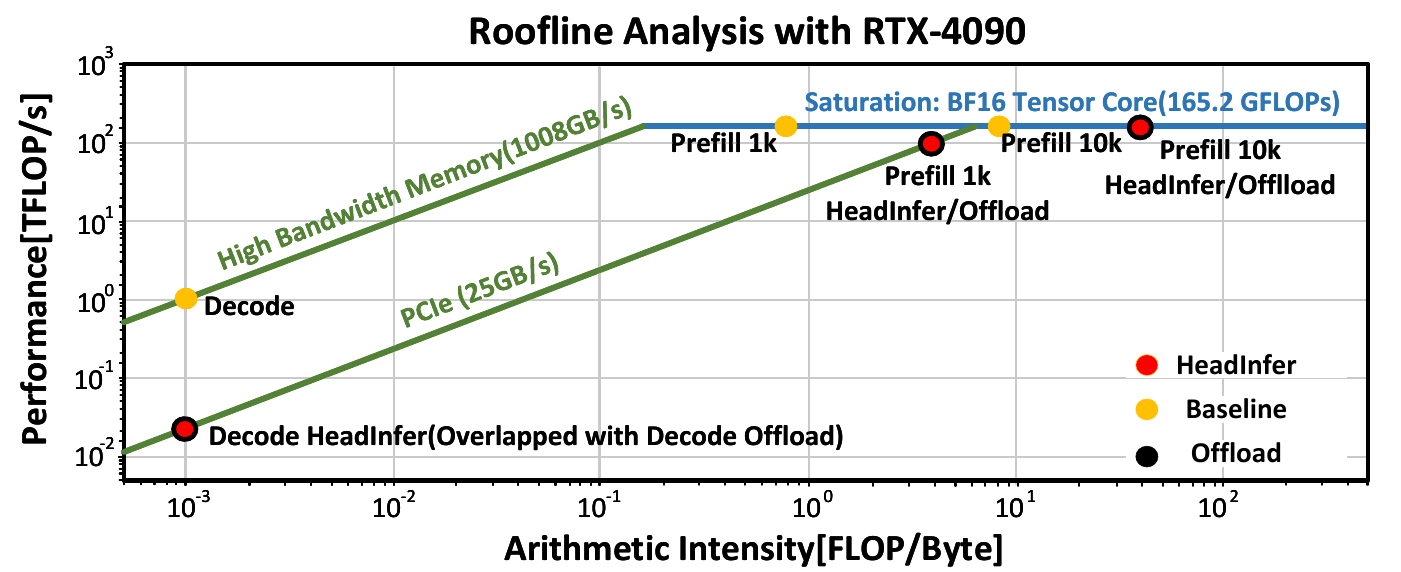}
  \vspace{-1 em}
  \caption{Flashattention in the roofline plot analysis using the RTX-4090 device setting.}
  \label{fig:roofline}
  \vspace{-2em}
\end{center}
\end{figure*}

\definecolor{comment_color_2}{RGB}{64,128,128}
\newcommand{\LineComment}[1]{\vspace*{0.5em}\small\textcolor{comment_color_2}{\textit{\# #1}}}
\restylefloat{algorithm}

\vspace{-1em}
\begin{algorithm}[H]
\caption{\sysname: Head-Wise Inference}
\label{alg:headinfer}
\begin{algorithmic}[1]
\small
\REQUIRE Transformer model with $L$ layers, $H$ attention heads per layer, input sequence length $S$ \\
\LineComment{  \textbf{Initialization:}}
\STATE Allocate GPU memory for $H_\text{on}$ heads.
\STATE Allocate CPU memory for $H_\text{off}$ heads.

\LineComment{ \textbf{Prefill Phase:}}
\FOR{each chunk of input tokens $S$}
    \FOR{each layer $l$ in Transformer Layers $L$}
        \FOR{each head $h$ in $H$}{}
            \STATE Compute key $K^{(h)}$, value $V^{(h)}$ on GPU
            \IF{$h \in H_\text{on}$ (GPU-resident heads)}
                \STATE Update GPU KV cache: $K_\text{GPU}^{(h)}$, $V_\text{GPU}^{(h)}$
            \ELSE                
                \STATE Async Prefetch next KV cache $K^{(h+1)}$, $V^{(h+1)}$
                \STATE Async Update CPU KV cache: $K_\text{CPU}^{(h)}$, $V_\text{CPU}^{(h)}$
            \ENDIF
            \STATE Compute attention output $A_t^{(h)}$ on GPU
        \ENDFOR
        \STATE Concatenate $A_t^{(h)}$ for all heads to form layer output
    \ENDFOR
    
\ENDFOR
\\
\LineComment{ \textbf{Decoding Phase:}}
\FOR{each token $t$ to be generated}
    \FOR{each layer $l$ in Transformer Layers $L$}
        \FOR{each head $h$ in $H$}
            \STATE Compute key $K^{(h)}$, value $V^{(h)}$ on GPU
            \IF{$h \in H_\text{on}$ (GPU-resident heads)}
                \STATE Update GPU KV cache: $K_\text{GPU}^{(h)}$, $V_\text{GPU}^{(h)}$
            \ELSE
                \STATE Async Prefetch next KV cache $K_\text{CPU}^{(h)}$, $V_\text{CPU}^{(h)}$
                \STATE Async Updated CPU KV cache: $K_\text{CPU}^{(h)}$, $V_\text{CPU}^{(h)}$
            \ENDIF
            \STATE Compute attention output $A_t^{(h)}$ on GPU
        \ENDFOR
        \STATE Concatenate $A_t^{(h)}$ for all heads to form layer output
    \ENDFOR
\ENDFOR

\end{algorithmic}
\end{algorithm}

\section{Analysis }
\label{Analysis}
Although KV cache offload is proposed to reduce memory usage, it remains an open question whether offloading harms overall performance, especially when the context length $S$ is large and works with chunked-prefill. This section analyzes the theoretical peak performance for a given GPU under constrained high-bandwidth memory (HBM) and peripheral component interconnect express (PCIe).

\textbf{Performance Model.} We consider a GPU characterized by HBM capacity $M_{HBM}$, memory bandwidth $B_{mem}$, and compute throughput $F_{GPU}$ (measured in FLOPS). We also incorporate the slower PCIe bandwidth $B_{pcie}$ into the performance model to account for the offload.  

\textbf{Memory Bound vs. Compute Bound.} GPU operators can be classified as compute bound or memory bound, which is determined by the time spent in arithmetic operations and the time spent accessing HBM. Two primary bottlenecks define the system’s performance regime:

\begin{itemize}
\item Memory-Bound: When $B_{mem}$ (memory bandwidth) is insufficient to transfer the KV cache quickly enough, inference operates below the GPU’s peak FLOPS capacity.

\item Compute-Bound: When $B_{mem}$ is sufficient, high compute efficiency is achieved and the throughput is determined by the GPU’s peak computation rate $F_{GPU}$.

\end{itemize}

\textbf{Peak Performance.}
Assume, for simplicity, that we can approximate the per-token inference time $T$ as follows:
\begin{equation}
 T \approx max(T_{comp}, T_{mem})
\end{equation}
During the prefill stage, $T_{comp} \propto \frac{D\times S^2 \times L \times H}{F_{GPU}}$ captures the compute part, and $T_{mem} \propto \frac{D\times S \times L \times H}{B_{mem}}$ captures the memory part. When scaling $S$, $T_{comp}$ grows faster than $T_{mem}$, associated with the quadratic growth in sequence length, which makes compute throughput the limiting factor.

During the decoding stage, each new token must attend to all previously generated tokens. Generating a single token incurs costs is roughly: $T_{comp} \propto \frac{D\times S \times L \times H}{F_{GPU}}$ and $T_{mem} \propto \frac{D\times S \times L \times H}{B_{mem}}$. When scaling $S$, $T_{comp}$ grows linearly as $T_{mem}$, but $B_{mem}, B_{pcie}$ are usually slower than $F_{GPU}$, making the memory bandwidth the main bottleneck.

\begin{table*}[ht]
\centering
\vspace{-0.5em}
\caption{Performance(benchmark score) of different methods on \textbf{LongBench v2} on a single RTX-4090 GPU, under different task difficulties (Easy/Hard) and context lengths (Short/Medium/Long). Overall performance is average scores on all questions.}
\vspace{0.5em}
\label{tab:longbenchv2}

\begin{tabular}{l|cccccc}
\hline
LongBench V2 & Overall & Easy & Hard & Short & Medium & Long \\
\hline
Standard 25K & 28.4 & 30.2 & 27.3 & 33.9 & 25.1 & 25.9 \\
Chunked Prefill 30K & 28.2 & 27.1 & 28.9 & 32.8 & 25.6 & 25.9 \\
Layer-wise offload 45K & 29.0 & 29.2 & 28.9 & \textbf{36.1} & 24.2 & 26.9 \\
\textbf{\sysname 1024K} & \textbf{30.2} & \textbf{31.2} & \textbf{29.6} & 33.9 & \textbf{27.0} & \textbf{30.6} \\
\hline
\end{tabular}
\vspace{-1em}
\end{table*}

\begin{table*}[h]
\centering
\vspace{-0.5em}
\caption{Performance(benchmark score) of different methods on \textbf{SCBench} on a single RTX-4090 GPU. kv and prefix-suffix are string retrieval in key-value and prefix-suffix scenarios. vt is variable tracking. qa-chn, qa-eng, and choice-eng are English/Chinese question answering. mf is finding the math answer. many-shot is finding multiple shots in context. summary is document summarization. }
\vspace{0.5em}
\begin{tabular}{l|ccccccccc}
\hline
SCBench & kv & prefix-suffix & vt & qa-chn & qa-eng & choice-eng & mf & many-shot & summary \\
\hline
Standard 25K & 15.8 & 9.6 & 4.6 & 9.4 & 13.3 & 36.5 & 2.6 & 16.3 & 32.3 \\
Chunked Prefill 30K & 21.4 & 10.4 & 6.9 & 9.4 & 15.5 & 38.6 & 2.2 & 25.2 & 33.5 \\
Layer-wise offload 45K & 22.6 & 12.8 & 8.4 & 10.4 & 15.7 & 37.8 & 2.2 & 25.9 & 33.6 \\
\textbf{\sysname 1024K} & \textbf{28} & \textbf{17.2} & \textbf{42} & \textbf{11.9} & \textbf{23.0} & \textbf{59.8} & \textbf{9.4} & \textbf{25.9} & \textbf{37.1} \\
\hline
\end{tabular}
\vspace{-1em}
\label{tab:llama3-comparison}
\end{table*}

\textbf{Roofline Model.} The roofline plot typically displays a kernel's computation throughput and arithmetic intensity, providing a visual representation of its hardware performance. We present the roofline analysis for the FlashAttention kernel \cite{dao2022flashattention,dao2023flashattention,shah2024flashattention} executed on an RTX-4090 (Figure \ref{fig:roofline}), with details in Appendix \ref{Rooflineappendix}. We believe that the roofline model can analyze both GPU performance and heterogeneous systems performance.

Our key observations from roofline analysis are:

\vspace{-1em}
\begin{itemize}
\item \textbf{Prefill (compute-bound behavior).} "Prefill \sysname /Offload" has higher arithmetic intensity than "Prefill" as it only offloads KV cache, and the arithmetic intensity grows as context length increases. For context lengths $S\geq10\mathrm{k}$, prefill remains compute-bound, even when offloading via slower PCIe bandwidth. In contrast, for short context lengths ($S\leq1\mathrm{k}$), “\sysname /Offload” shifts to the memory-bound regime, leading to offload-induced slowdowns. The turning point is achieved at 2K, and 10K can ensure high computational efficiency.


\item \textbf{Decoding (memory-bound behavior).} Decoding performance is primarily memory-bound. Consequently, relying on PCIe bandwidth during offloading substantially degrades overall throughput.

\item \textbf{\sysname (unchanged behavior).} \sysname's head-wise offloading during both chunked-prefill and decoding does not alter the position of the roofline plot due to the independence of the attention heads. Although head-wise computation impacts GPU utilization, the adaptive head strategy can preserve performance.
\end{itemize}

\section{Performance Evaluation}

\subsection{Experimental Setup}
We conduct our experiments on RTX-4090 GPUs, which are configured with 24.5 GB of HBM3, 4× AMD EPYC 7B13 with 64 CPUs each (a total of 256 cores), and 2× Gen4 NVMe of 512 GB each and 1 TB of DDR5 RAM in total. The GPUs are independently connected to the host with 16 PCIe 4.0 interfaces, providing 25.0 GB/s unidirectional D2H and H2D throughput for pinned host memory.

\subsection{Long-Context Benchmarks}
We evaluate \sysname using the LongBench v2 \cite{bai2024longbench2} and SCBench \cite{li2024scbench} benchmarks, and other long-context benchmark results such as Needle-in-a-Haystack (NIAH) \cite{kamradt2023needle} and Ruler \cite{hsieh2024ruler} are shown in the Appendix \ref{Long-Context Benchmarks Details}. We use the Llama-3-8B-Instruct-Gradient-1024k model, which supports up to 1 million context lengths. All lossless methods, including standard inference, chunked prefill, layer-wise offload, and \sysname, are running on a single RTX-4090 GPU, with a maximal context length achieved within 24 GB.  The long-context benchmark results aim to reveal the performance gain with \sysname's context length extensions.

\begin{table*}[h]
\centering
\vspace{-0.5em}
\caption{Comparison of maximum context length with system optimization methods on various models inference. All experiments within this table run on a single RTX-4090 GPU with 24GB and AMD EPYC 7V12 CPUs with 1TB memory.}
\vspace{0.5em}
\begin{tabular}{lccccc}
\toprule
Context Length & Llama-3-8B & Llama-2-7B & Mistral-7B & Qwen2-7B & Gemma-2-9b \\
\midrule
Standard & 25K & 10K & 30K & 35K & 10K \\
\midrule
Chunked Prefill & 30K & 20K & 40K & 70K & 10K \\
4-bit KV-quant & 45K & 30K & 40K & 50K & 20K \\
Layer-wise offload & 45K & 60K & 45K & 50K & 35K \\
\textbf{\sysname} & \textbf{4096K} & \textbf{1024K} & \textbf{4096K} & \textbf{4200K} & \textbf{1300K} \\
\hline
\end{tabular}

\vspace{-2em}
\label{tab:all-comparison}
\end{table*}

\begin{table}[h]
\centering
\caption{ Llama3-70B Inference with long context input.}
\vspace{0.5em}
\begin{tabular}{lc}
\toprule
Context Length & Llama-3-70B \\
\midrule
Standard & 10K \\
\midrule
\textbf{\sysname + 10k chunk size} & \textbf{950K} \\
\textbf{\sysname + 5k chunk size} & \textbf{1000K} \\
\hline
\end{tabular}

\vspace{-1em}
\label{tab:70B-comparison}
\end{table}

\textbf{LongBench v2} is a comprehensive suite of long-context datasets designed to assess long-context problems that require deep understanding and reasoning. It comprises 503 difficult multiple-choice questions within two difficulties, including "Easy/Hard" and word lengths, including "Short" of 0-32k words, "Medium" of 32k-128k words, and "Long" of 128k-2M words. As shown in Table \ref{tab:longbenchv2}, we measure their overall scores for different task categories. \sysname achieves the highest benchmark score for "Medium" (27.70) and "Long" (30.6). Layer-wise offload performs best on "Short" (36.1) for truncation side effects. 

\textbf{SCBench} is a comprehensive suite of datasets encompassing multiple tasks, designed to assess long-context understanding capabilities. It contains subsets with different prompt lengths up to 500K tokens and 227K on average. As shown in Table \ref{tab:llama3-comparison}, we measure their overall scores for different task categories. \sysname outperforms all other methods on all 9 tasks, with superior 1M context length.

\subsection{Efficiency Results of Memory and Throughput}

We evaluate the maximum context length supported under RTX-4090 memory constraints, as well as prefill/decoding throughput. Our experiments use the  Llama-3-8B, Llama-3-70B, Mistral \cite{jiang2023mistral}, Qwen \cite{bai2023qwen}, and Gemma-2 \cite{team2024gemma} models. The default number format for weights and activations is BFloat16, and 4-bit KV-quant is deployed with KIVI \cite{liu2024kivi}. The chunk size is set to 10K based on our roofline analysis.

\textbf{LLM Inference on Consumer GPUs with 24G memory.} We measure the GPU memory consumption of Llama3-8B inference with \sysname and 1 million context length. \sysname uses 17GB during prefill and 16.4GB during decoding; in contrast, other methods are unable to run at this context scale using 24GB RTX-4090. Accordingly, we measure the maximum achievable context length to assess memory efficiency. As shown in Table \ref{tab:all-comparison}, \sysname outperforms other system optimization approaches, scaling from thousand-level contexts (10K-70K) to million-level contexts (1,024K–4,200K). Compared to layer-wise offload, \sysname can maintain stable activation memory with chunked prefill and minimize KV cache GPU memory with head-wise offload. Note that with \sysname, the maximum context lengths for Llama3, Llama2, and Mistral are constrained by CPU RAM (512GB for KV cache), while the other methods are limited by GPU memory. This means we can use larger CPU RAM or offload to disk for a more extended context. We leave this exploration for future work.

\textbf{Scaling up to Llama3 70B Architecture.} Scaling ability to 70B models is a key factor for demonstrating if \sysname is effective for larger-scale model inference. We evaluate \sysname on the Llama-3-70B model. The model inference is conducted using pipeline parallelism \cite{narayanan2019pipedream} across 8 RTX-4090 GPU nodes. As shown in table \ref{tab:70B-comparison}, \sysname achieves efficient inference of 1 million sequences with the 70B model, outperforming standard methods. Additionally, due to Llama-3-70B’s higher arithmetic intensity compared to Llama-3-8B, we can employ a smaller chunk size without performance degradation. This reduced chunk size allows for better memory utilization, enabling increased KV cache allocation and extending the maximum context length from 950k to 1M.


\textbf{Prefill and Decoding Throughput of \sysname}
We evaluate the prefill and decode throughput of the inference Llama-3-8B model using \sysname with adaptive head grouping. Our \sysname implementation achieves 516 tokens/second during prefill for a 1 million-token input and 0.15 tokens/second during decoding with a 1 million-token KV cache. A context length of 1 million cannot be achieved by other methods. For comparison, with a 20K-token input, \sysname achieves 7210 tokens/second prefill and 6 tokens/second decoding. Standard inference achieves 7235 tokens/second prefill and 33 tokens/second decoding, while layer-wise offloading matches \sysname at 7210 tokens/second prefill and 6 tokens/second decoding. Additional latency results for different context lengths and different methods are provided in Appendix \ref{latency}.

\section{Ablation Study}

\textbf{How do different levels of granularity affect memory?}
To evaluate the impact of different levels of granularity on memory, we conducted a detailed ablation study using the Llama-3-8B model. From the memory perspective, three dimensions of sequences $S$, layer $L$, and head $H$ are managed through chunked-prefill, layer-wise offloading, and head-wise offloading, respectively. 
As shown in Table \ref{tab:Ablation Study}, \sysname with $S \times L \times H$ supports a 4096K context length, far exceeding the 25K of standard methods, 45K of layer-wise offload with $L$, and chunked prefill with $S$. Adjusting head-group granularity shows a trade-off, with context lengths ranging from 2100K to 550K. Note that \sysname with (Head=8 Group=1) works the same as layer-wise offloading combined with chunked-prefill for $S \times L$. \sysname achieves $8 \times$ context extension than (Head=8 Group=1), showing its memory efficiency.


\begin{table}[t]
\centering
\caption{ Ablation study of \sysname on Llama-3-8B.}
\vspace{0.5em}
\begin{tabular}{lc}
\toprule
Context Length & Llama-3-8B \\
\midrule
Standard & 25K \\
\midrule
Layer-wise Offload & 45K \\
Chunked Prefill & 30K \\
\sysname Head=8 Group = 1 & 550K \\
\sysname Head=4 Group = 2 & 1100K \\
\sysname Head=2 Group = 4 & 2100K \\
\textbf{\sysname} & \textbf{4096K} \\
\hline
\end{tabular}

\vspace{-2em}
\label{tab:Ablation Study}
\end{table}

\section{Conclusion}
In this paper, we introduced \sysname, a novel head-wise KV cache management framework designed to enable efficient long-context large language model (LLM) inference on consumer GPUs. \sysname dynamically offloads KV cache components to CPU memory, employing head-wise and asynchronous offloading strategies. Our Roofline analysis highlights \sysname's ability to preserve GPU computational efficiency while reducing memory footprint, making long-context inference feasible on consumer-grade GPUs. Our evaluations demonstrated \sysname's ability to achieve over 1-4 million context tokens on a single consumer GPU with mathematical equivalency.

We hope that our work will inspire future research in memory-efficient inference from the perspective of head-wise offload. We believe that \sysname will be a valuable tool for the community, enabling the inference of long context on consumer-grade hardware with limited resources.

\section*{Impact Statement: Democratizing Access to Advanced AI}

Artificial Intelligence (AI) has the potential to transform industries, revolutionize education, and empower individuals. However, the deployment of cutting-edge models, particularly large language models (LLMs), is often hindered by significant resource requirements, creating barriers to entry for smaller organizations and underserved communities.

In this work, we introduce \sysname, a memory-efficient inference framework designed to bridge this gap. By leveraging head-wise offloading strategies, \sysname enables resource-constrained devices to process unprecedentedly long contexts, achieving capabilities typically reserved for high-performance systems. For instance, \sysname allows a consumer-grade GPU to handle over 1 million context tokens, democratizing access to advanced LLM functionalities.


Ultimately, \sysname aligns with the broader vision of democratizing AI, ensuring that technological advancements benefit humanity as a whole rather than a select few. Through innovations like \sysname, we hope to inspire a future where AI is universally accessible, fostering creativity, knowledge sharing, and inclusive progress.

\nocite{langley00}

\bibliography{example_paper}
\bibliographystyle{icml2025}

\newpage
\appendix
\onecolumn

\section{Experiment Details}
\subsection{Prefill and Decoding Overhead of \sysname}
\label{latency}
We evaluate the computational overhead of \sysname compared to baseline approaches across different context lengths using the Llama-3-8B model. Our analysis focuses on two key phases: prefill and decoding.

\begin{table}[H]
\centering
\caption{Prefill overhead (in seconds) of Llama3-8B under different context lengths.}
\label{tab:prefill-overhead}
\begin{tabular}{lcccccccccc}
\toprule
\textbf{Prefill Latency(s)} & \textbf{1K} & \textbf{10K} & \textbf{20K} & \textbf{40K} & \textbf{100K} & \textbf{200K} & \textbf{400K} & \textbf{1M} & \textbf{2M} & \textbf{4M} \\
\midrule
Standard        & 0.11        & 1.23         & 2.83         & -            & -             & -             & -             & -           & -           & -           \\
Chunked Prefill           & 0.11        & 1.23         & 2.83         & -            & -             & -             & -             & -           & -           & -           \\
Layer-offload   & 0.12        & 1.24         & 2.84         & 6.93         & -             & -             & -             & -           & -           & -           \\
\sysname (head=8/group=1) & 0.12        & 1.24         & 2.84         & 7.11         & 30.2          & 100           & 357           & -           & -           & -           \\
\sysname (head=4/group=2) & 0.13        & 1.23         & 2.89         & 7.26         & 30.2          & 99            & 351           & 2033        & -           & -           \\
\sysname (head=2/group=4) & 0.14        & 1.23         & 2.94         & 7.54         & 30.5          & 100           & 353           & 2035        & 7952        & -           \\
\sysname (head=1/group=8) & 0.21        & 1.27         & 3.06         & 7.77         & 31.2          & 101           & 356           & 2054        & 7975        & 27114       \\
\sysname Adaptive        & 0.13        & 1.24         & 2.84         & 7.11         & 30.2          & 99            & 351           & 2033        & 7952        & 27114       \\
\bottomrule
\end{tabular}
\end{table}

\begin{table}[H]
\centering
\caption{Decoding overhead (in seconds per generated token) of Llama3-8B under different KV cache (context) sizes.}
\label{tab:decode-overhead}
\begin{tabular}{lcccccccccc}
\toprule
\textbf{Decoding Latency(s)} & \textbf{1K} & \textbf{10K} & \textbf{20K} & \textbf{40K} & \textbf{100K} & \textbf{200K} & \textbf{400K} & \textbf{1M} & \textbf{2M} & \textbf{4M} \\
\midrule
Standard        & 0.03        & 0.03         & 0.03         & 0.04         & -             & -             & -             & -           & -           & -           \\
Chunked Prefill           & 0.03        & 0.03         & 0.03         & 0.04         & -             & -             & -             & -           & -           & -           \\
Layer-offload   & 0.03        & 0.09         & 0.17         & 0.28         & 0.66          & 1.3          & 2.58          & -           & -           & -           \\
\sysname (head=8/group=1) & 0.03        & 0.09         & 0.17         & 0.28         & 0.66          & 1.3          & 2.58          & -           & -           & -           \\
\sysname (head=4/group=2) & 0.04        & 0.10         & 0.16         & 0.28         & 0.67          & 1.31          & 2.58          & 6.41        & -           & -           \\
\sysname (head=2/group=4) & 0.06        & 0.11         & 0.17         & 0.30         & 0.68          & 1.32          & 2.59          & 6.46        & 13.7        & -           \\
\sysname (head=1/group=8) & 0.10        & 0.14         & 0.21         & 0.33         & 0.71          & 1.33          & 2.61          & 6.51        & 13.8        & 27.2        \\
\sysname Adaptive        & 0.03        & 0.09         & 0.17         & 0.28         & 0.66          & 1.73          & 3.03          & 6.41        & 13.7        & 27.2        \\
\bottomrule
\end{tabular}
\end{table}

For prefill operations, \sysname demonstrates similar performance to standard approaches for shorter context lengths (up to 20K tokens). Beyond this range, \sysname scales efficiently with longer contexts, outperforming Layer-offload due to its fine-grained head-wise KV cache management. Notably, \sysname enables inference for 4M tokens on a single GPU, which is otherwise infeasible.

In decoding, \sysname maintains low latency at short and medium context lengths. For extended contexts (e.g., 1M and beyond), \sysname introduces manageable latency while supporting unprecedented context lengths, with adaptive configurations optimizing performance further.

The performance of \sysname relies on its ability to dynamically adapt to varying hardware constraints and workload requirements by controlling the granularity of offloading. Specifically, HeadInfer Adaptive achieves optimal performance by selectively choosing the most suitable head group size based on the context length and memory limitations.

\textbf{HeadInfer (Head = 8 / Group = 1).} This configuration aggregates all attention heads within a layer into a single group for offloading, effectively making the KV cache management layer-wise. As such, \textbf{HeadInfer (Head Group = 8)} is functionally equivalent to \textbf{Layer-Offload}, where all KV cache for a layer is stored on either the GPU or offloaded to the CPU in a single operation. However, HeadInfer provides the flexibility to adjust the granularity of KV cache management beyond this layer-wise approach.

\textbf{HeadInfer (Head = 1 / Group = 8).} At the other extreme, this configuration offloads each attention head individually, offering the finest level of granularity for KV cache control. While this achieves the highest memory savings, frequent PCIe transfers and kernel launches introduce overhead, especially for shorter context lengths, which can impact throughput.

\textbf{HeadInfer Adaptive} dynamically selects the optimal head group size ($h$) to balance memory efficiency and computational throughput. This adaptability allows it to achieve superior performance across diverse hardware and context length requirements:
\begin{itemize}
    \item \textbf{Shorter Contexts:} For shorter context lengths, PCIe transfer and kernel launch overheads dominate. Larger head groups (e.g., $h = 4$ or $h = 8$) are preferred to minimize overhead while maintaining memory efficiency.
    \item \textbf{Longer Contexts:} For longer context lengths, memory usage becomes the primary bottleneck. Smaller head groups (e.g., $h = 2$ or $h = 1$) are chosen to maximize context length while reducing GPU memory usage.
\end{itemize}
This dynamic strategy enables \textbf{HeadInfer Adaptive} to consistently deliver the \textbf{best performance} by adapting to both memory and compute-bound regimes, as observed in performance and overhead evaluations.

In summary:
\begin{itemize}
    \item \textbf{HeadInfer Adaptive} balances memory usage and performance by dynamically adjusting head group sizes, providing the \textbf{best of both worlds}: memory efficiency and scalability.
    \item \textbf{HeadInfer (Head Group = 8)} is equivalent to \textbf{Layer-Offload}, serving as a robust baseline for layer-level KV cache management.
    \item The head-wise approach of \sysname enables \textbf{extended context lengths during prefill}, maintaining high throughput without requiring specialized hardware or approximation techniques.
\end{itemize}

This adaptability positions \textbf{HeadInfer Adaptive} as an essential method for large-scale language model inference in memory-constrained environments.

\subsection{Long-Context Benchmarks Details}
\label{Long-Context Benchmarks Details}

\begin{figure*}[t]

\centering
\begin{minipage}[c]{0.5\textwidth}
  \centering
  \includegraphics[width=\linewidth]{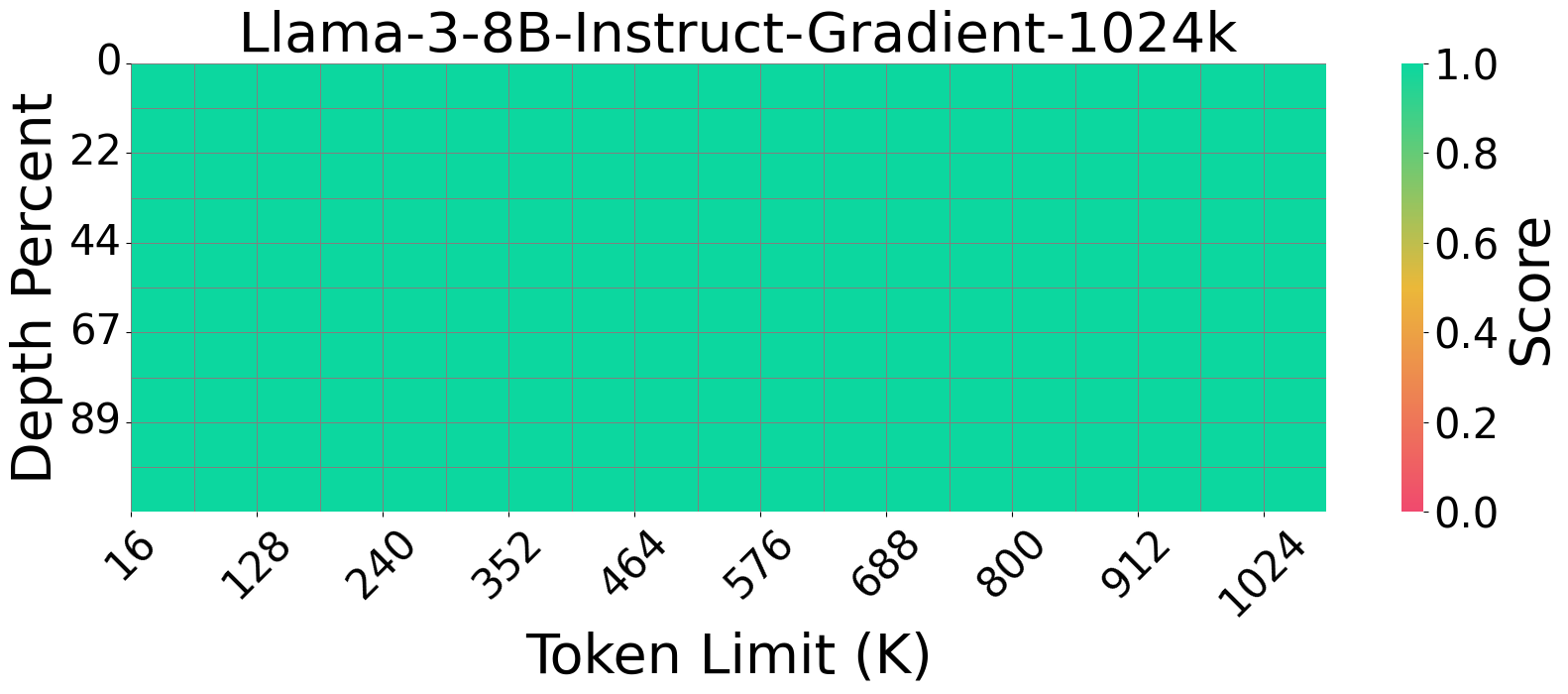}
\end{minipage}
\vspace{-0.5em}
\caption{\sysname provides equal accuracy as standard inference on the Needle-in-a-Haystack benchmark}\label{fig:Needle}

\vspace{-1em}
\end{figure*}

\textbf{Needle-in-a-Haystack \cite{kamradt2023needle}}  is a challenging pressure test designed to assess the ability of models
to accurately identify and retrieve relevant information from a lengthy context.
Figure \ref{fig:Needle} presents the results on Needle In A Haystack. We use 1024K for Llama-3-8B-Instruct-Gradient-1024k. \sysname can accurately recall the information queried within the context in question. 

\textbf{Ruler \cite{hsieh2024ruler}.} Designed for a comprehensive evaluation of long context, the Ruler benchmark is a recent synthetic benchmark suite that includes 13 complex tasks in four main categories. Each context length variation includes 2,600 examples, with tests conducted at 4K, 8K, 16K, 32K, 64K, and 128K tokens. The benchmark comprises four key categories. The retrieval category includes various Needle-in-a-Haystack tasks: Single (S-NIAH) for finding individual key-value pairs in noisy text, Multi-keys (MK-NIAH) for retrieving specific values among hard distractors, Multi-values (MV-NIAH) for finding all values linked to a single key, and Multi-queries (MQ-NIAH) for retrieving values across multiple keys. The Multi-hop Tracing category features Variable Tracking (VT), requiring models to trace and return all variable names pointing to the same value through variable bindings. For aggregation, the benchmark includes Common Words Extraction (CWE) for identifying top-K common words from mixed sets and Frequent Words Extraction (FWE) for finding the most frequent words from a Zeta distribution. The Question Answering category extends traditional QA datasets by adding distracting paragraphs, testing models' ability to locate and utilize relevant information amid distractors.

\begin{table}[h]
\centering
\caption{Maximum achievable sequence lengths for different inference methods}
\begin{tabular}{lc}
\toprule
Method & Supported Sequence Length within Ruler \\
\midrule
Standard Inference & 16K \\
Chunked-Prefill & 32K \\
Layer-wise Offload & 32K \\
\sysname & 128K \\
\bottomrule
\end{tabular}
\label{ruler_context}
\end{table}

Table \ref{ruler_context} demonstrates the maximum achievable sequence lengths for different inference methods on the Ruler benchmark. Standard inference, while straightforward, is limited to 16K tokens due to memory constraints. Both Chunked-Prefill and Layer-wise Offload methods double this capacity to 32K tokens through their respective optimization strategies. \sysname shows a significant advancement by enabling processing of sequences up to 128K tokens - a 4x improvement over other offloading methods and a 8x improvement over standard inference. This extension in sequence length is achieved through \sysname's novel head-wise offloading strategy, which more efficiently manages GPU memory utilization.

\begin{table}[h]
\centering
\caption{Performance on Ruler benchmark tasks across different context lengths}
\small
\begin{tabular}{lcccccccc}
\toprule
Context & NIAH & MK-2 & MK-3 & VT & CWE & FWE & QA-1 & QA-2 \\
Length & (\%) & (\%) & (\%) & (\%) & (\%) & (\%) & (\%) & (\%) \\
\midrule
4K  & 100.0 & 99.6 & 100.0 & 99.20 & 99.38 & 94.53 & 84.6 & 59.8 \\
8K  & 100.0 & 99.8 & 99.6  & 99.08 & 94.68 & 84.93 & 79.2 & 56.2 \\
16K & 100.0 & 100.0 & 99.4 & 98.72 & 56.90 & 90.60 & 79.6 & 53.2 \\
32K & 100.0 & 99.6 & 99.8  & 97.32 & 2.78  & 93.20 & 77.2 & 50.4 \\
64K\textsuperscript{} & 100.0 & 97.4 & 97.8  & 92.48 & 0.10  & 84.27 & 76.0 & 49.4 \\
128K\textsuperscript{} & 100.0 & 75.2 & 56.6  & 54.68 & 0.10  & 74.8 & 71.8 & 41.2 \\
\bottomrule
\multicolumn{9}{l}{\textsuperscript{}\sysname only} \\
\end{tabular}
\label{ruler_results}
\end{table}

Table \ref{ruler_results} presents a comprehensive analysis of performance across different tasks and context lengths. Since the ruler designs different tasks for different context lengths, the ruler's Multi-hop Tracing, Aggregation, and Question Answering become more difficult to complete in long sequence scenarios. This is different from scbench, which directly truncates long sequence tasks. However, even in this case, \sysname can still show the same performance as standard inference, which confirms its mathematical consistency.


\subsection{Memory Analysis of Llama-3-8B Inference}
\label{Memory Analysis of Inference LLama3-8B}

In this work, we compare several inference configurations, each targeting different trade-offs between memory efficiency and inference speed. While Figure \ref{fig:first_analysis} gives the theoretical benefit of \sysname compared to other methods in terms of memory usage, we also give how the theoretical memory is calculated here. Table \ref{tab:memory_analysis_fig1} provides a comparative breakdown of memory usage across different inference strategies. These methods balance GPU memory consumption for weights, KV-cache, and activations while addressing the challenges of scaling to long-context inputs. Each strategy is outlined below:

\begin{itemize}
    \item \textbf{Standard}: The standard inference method keeps all weights, KV-cache, and activations entirely in GPU memory. With a context length of $S$, the KV-cache scales as $S \times L \times D_h \times H$, where $L$ is the number of layers, $D_h$ is the hidden dimension per head, and $H$ is the number of attention heads. Activations require additional memory of $S \times D + 2 \times S \times I$, where $I$ is the intermediate MLP dimension. While this approach achieves baseline performance, it is constrained by GPU memory limits (e.g., 128 GB for the KV-cache and 207GB for toal).

    \item \textbf{Chunked-Prefill}: By dividing the input into smaller chunks of size \texttt{chunk}, this method reduces activation memory from $S$ to \texttt{chunk}. The memory footprint for activations is \texttt{chunk}$\times D + 2 \times$\texttt{chunk}$\times I$, where only part of the sequence resides in GPU memory during processing. Although the KV-cache size remains $S \times L \times D_h \times H$, this technique significantly lowers activation memory requirements, reduce total memory from 207GB to 143GB.

    \item \textbf{4-bit KV-Quant}: This method compresses the KV-cache from fp16/bf16 (16-bit floating point) to a 4-bit representation, reducing its size by a factor of 4. The memory usage becomes $S \times L \times D_h \times H / 4$, while activations remain $S \times D + 2 \times S \times I$.

    \item \textbf{Layer-wise Offload}: This strategy offloads the KV-cache for entire layers to CPU memory as soon as computation for that layer is complete. On GPU, the memory required for KV-cache is reduced to $S \times D_h \times H \times 2$, the final $2$ due to the ping-pong memory mechanism. However, offloading incurs communication overhead, making this approach more suitable for scenarios with sufficient PCIe bandwidth.

    \item \textbf{\sysname}: Our proposed approach achieves fine-grained control over memory by offloading KV-cache at the attention head level. With a ping-pong memory mechanism, the on-GPU KV-cache is reduced to $S \times D_h \times 2$, significantly lowering memory requirements. Activations are further minimized by combining chunked prefill with selective offloading. \sysname enables unprecedented scaling to 1M tokens and beyond, allowing context lengths of up to 4 million tokens with minimal GPU memory usage. Also, we observe the total estimated memory of 16.7 GB is close to the real measured memory of 17.0 GB, demonstrating the effectiveness of our estimation.
\end{itemize}

\begin{table}
\centering
\caption{Memory usage comparison for Llama3-8B with 1 million context length}
\label{tab:memory_analysis_fig1}
\begin{tabular}{lccccc}
\toprule
Method & Weight & KV-cache & Activation & Total & Total KV cache $^*$ \\
\midrule
Standard & 15.08 & 128 & 64 & 207 & 128 \\
Chunked Prefill & 15.08 & 128 & 0.625 & 143 & 128 \\
4bit-KV-quant & 15.08 & 32 & 64 & 111 & 32 \\
layer-wise-offload & 15.08 & 8 & 64 & 87 & 128 \\
\sysname & 15.08 & 1 & 0.625 & 16.7 & 128 \\
\bottomrule
\multicolumn{6}{l}{\small $^*$Total KV cache includes both GPU and CPU memory for offloading methods}
\end{tabular}
\end{table}

Table \ref{tab:memory_analysis} summarizes the GPU memory usage for weights, KV cache, and activations, as well as the total usage for in-device experiment. A detailed inference setting for each model and a comprehensive memory estimation of each method (including key hyperparameters, chunk sizes, and offloading policies) are provided here. As shown in table \ref{tab:memory_analysis}, we focus on five representative strategies for Llama-3-8B inference under varying context lengths:

\begin{table}[h]
\caption{Memory consumption analysis for different inference methods (in GB) on Llama-3-8B}
\label{tab:memory_analysis}
\centering
\begin{tabular}{lcccccc}
\toprule
Method & Weights & KV cache & Activation & Total & Total KV cache $^*$ \\
\midrule
Standard-25K & 15.08 & 3.13 & 1.56 & 19.77 & 3.13 \\
Chunked-Prefill-30K & 15.08 & 3.75 & 0.63 & 19.46 & 3.75 \\
4bit-KV-Quant-45K & 15.08 & 1.41 & 2.81 & 19.30 & 1.41 \\
Layer-Wise-Offload-45K & 15.08 & 0.35 & 2.81 & 18.25 & 5.63 \\
\sysname-4000K & 15.08 & 3.91 & 0.63 & 19.61 & 500 \\
\bottomrule
\multicolumn{6}{l}{\small $^*$Total KV cache includes both GPU and CPU memory for offloading methods}
\end{tabular}
\end{table}

\begin{itemize}[leftmargin=*] 
\item \textbf{Standard-25K:} Baseline approach with unmodified inference. All model weights, activations, and the entire KV cache stay on the GPU. Context length is limited by GPU memory. 
\item \textbf{Chunked-Prefill-30K:} Splits the prompt into sequential chunks to reduce activation overhead. KV cache remains on GPU. We use a chunk size (e.g., 4K or 8K tokens) such that each partial forward pass does not exceed GPU capacity. 
\item \textbf{4bit-KV-Quant-45K:} Applies a 4-bit quantization technique to KV cache, shrinking cache size by approximately 4$\times$ compared to FP16/BF16. However, the method can introduce additional on-GPU overhead in activation or conversion operations. In our tests, we adopt a standard quantization library \cite{hooper2024kvquant} for uniform 4-bit KV representation.
\item \textbf{Layer-Wise-Offload-45K:} Offloads each layer’s KV cache to CPU memory as soon as possible. During inference, only the KV cache for the currently processing layer is kept on the GPU. Once attention computation for that layer completes, its KV cache is swapped out to the CPU. This approach significantly lowers on-GPU KV cache usage but may incur additional offload overhead at each layer boundary. 
\item \textbf{\sysname-4000K:} Our proposed head-wise offloading approach (\sysname) that partitions the KV cache by attention heads. While some heads remain fully on GPU, the cache for other heads (or tokens) are immediately offloaded to CPU memory. Despite a very large total KV cache (reported as ``500 GB'' in Table~\ref{tab:memory_analysis}), only a small fraction resides on the GPU at any time. This enables context lengths in the order of millions of tokens if sufficient CPU RAM is available. \end{itemize}

For total memory we define: \begin{equation*} M_{\text{total}} ;=; M_{\text{weights}} + M_{\text{KV cache}} + M_{\text{activation}}. \end{equation*} \noindent \textbf{Weights} (\emph{e.g.}, 15.08GB in Llama-3-8B) remain constant for all methods. The \textbf{KV cache} size grows with the number of tokens processed (sequence length) but can be reduced or offloaded depending on the method. \textbf{Activation} memory arises from forward-pass intermediate tensors (attention blocks, MLP layers, etc.) and is partially minimized by chunking or parallelization strategies.

In Table~\ref{tab:memory_analysis}, \emph{KV cache} and \emph{Activation} columns refer to the approximate GPU memory usage during inference; any data offloaded to CPU memory (or disk) is not included in these columns but is counted in \emph{Total KV cache} if applicable. For instance, \textbf{\sysname} has a \emph{GPU-side} KV cache usage of about 3.9GB at any instant, whereas the \emph{overall} KV cache across CPU and GPU is up to 500GB, enabling very long context windows.

\section{Motivation and Additional Related Work}
\label{sec:Motivation}

In this section, we first explain that the KV cache size becomes a critical issue for long-text generation in LLM inference, and it becomes more problematic when deploying modern offloading-based inference systems. We then discuss why the existing KV cache management methods cannot fundamentally address the problem in an offloading-based system.

\subsection{Memory Requirements for Inference}
This section characterizes the memory requirements for transformer inference. It can be categorized into two components: i) Model Parameters and ii) Activation memory primarily referring to KV cache. The memory requirement for model parameters primarily depends on the hidden dimension $D$, the number of attention heads, and the number of Transformer layers $L$. Nearly all
the parameters in a Transformer block come from linear layers within $L$ attention blocks,  $L$ multilayer perceptron (MLP) blocks, and one language modeling head (LM-head) block. Take Llama-3-8B as an example; the total parameters in a transformer-based model can be approximated as $14 \times L \times HD^2$ with 15GB of memory. The memory required for activation memory primarily consists of the KV cache, which depends on the model
architecture, batch size $B$, and sequence length $S$, and it can be pretty significant. The memory can be estimated as $2\times B\times S \times H \times D$. For instance, in the Llama-3-8B \cite{dubey2024llama} model architecture, serving with FP16 KV cache for 1 million tokens would require at least 207 GB of memory—exceeding the capacity of a single 80GB GPU.

\begin{figure*}[t]
\begin{center}
  \includegraphics[scale=0.5]{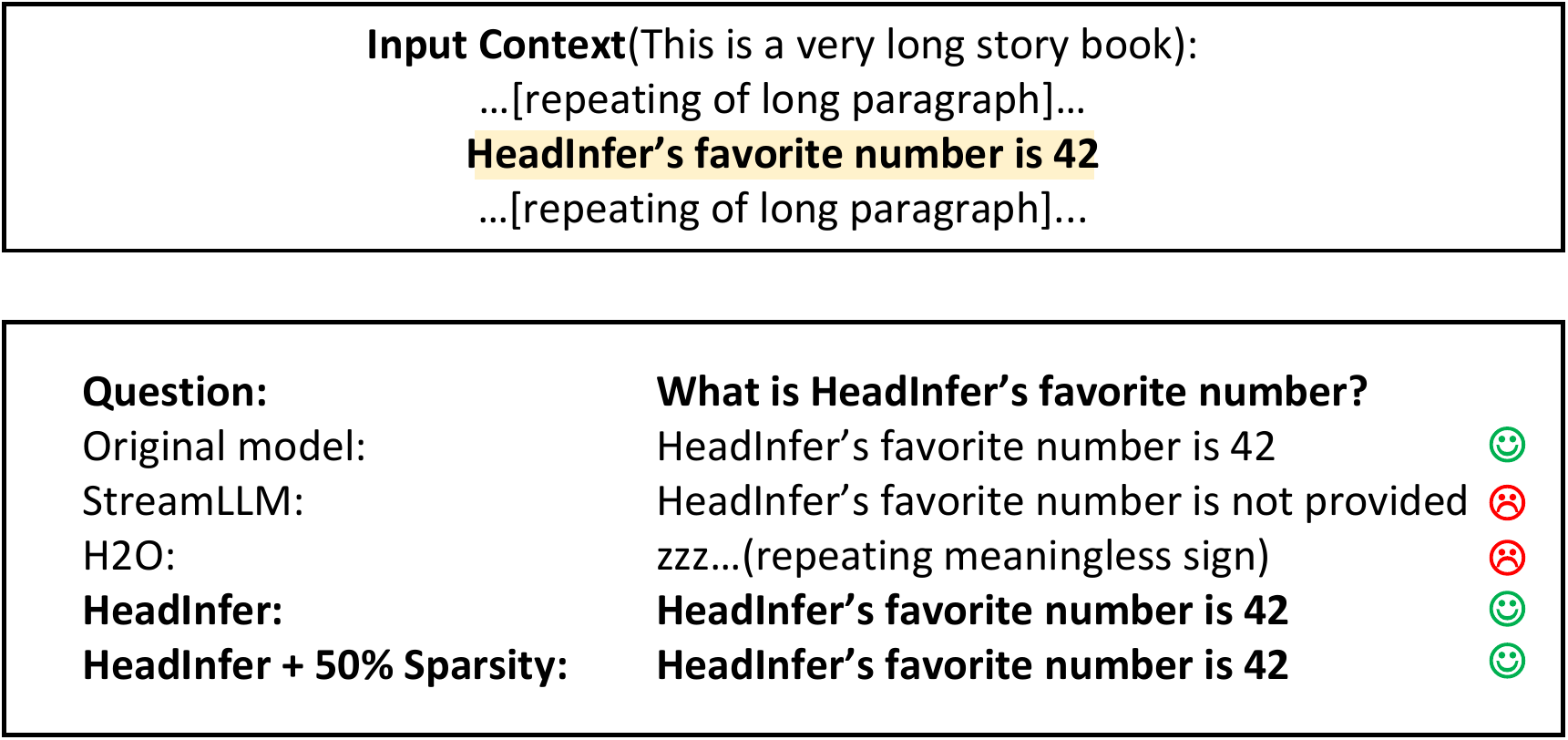}
  \caption{Token eviction methods cannot work when querying the less relevant information to the main theme. Here, we use a 10K document from LongBench \cite{bai2023longbench} and add one sentence that is not relevant to the main theme. In this case, H2O discards tokens less relevant to the main theme, leading to error generation. StreamingLLM discards tokens based on the query but remaining question tokens, making it Hallucinations. \sysname can successfully output the exact information from the lengthy input, even when we compress 75\% of the KV cache }
  \label{fig:input}
  \vspace{-1em}
\end{center}
\end{figure*}

\subsection{KV Cache in LLM Inference Systems}
As discussed in the previous section, today's LLM serving systems exploit KV caching to avoid redundant computation of key and value projections during the chunked-prefill decoding stage. While this is an effective solution for short sequence generation with a single client request, the KV cache quickly becomes a key memory consumer when we generate long sequences or employ modern request batching techniques \cite{sheng2023flexgen}.

In the Llama-3-8B \cite{dubey2024llama} model architecture, serving with FP16 KV cache for 1 million tokens would require at least 256 GB of memory—exceeding the capacity of a single 80GB GPU. Additionally, the latencies of pre-filling and decoding with such large contexts are significant, posing substantial challenges to the effective use of LLMs in long-context scenarios.

The rapidly expanding KV cache leads to an urgent need and numerous efforts for KV cache compression, particularly in scenarios with limited GPU memory. Architectural modifications, such as Grouped-Query Attention \cite{ainslie2023gqa}, Pose \cite{zhu2023pose}, Rope \cite{su2024roformer}, PI \cite{chen2023extending}, LongNet \cite{ding2023longnet}, MST \cite{luo2024mini}, LoQT \cite{loeschckeloqt}, Lora \cite{hu2021lora} and Galore \cite{zhao2024galore} require expensive model pre-training. One direction is non-Transformer architecture design, such as Mamba \cite{gu2023mamba}, Linear Attention \cite{katharopoulos2020transformers}, RWKV \cite{peng2023rwkv}, Griffin \cite{de2024griffin}. However, the transformer is still the most widely used model structure, and in this paper, we focus on KV cache reduction for typical transformers. KV cache token-drop methods, such as H2O \cite{zhang2024h2o}, StreamingLLM \cite{xiao2023efficient}, InfiniGen \cite{lee2024infinigen}, ClusterKV \cite{liu2024clusterkv} often compromise accuracy in long-context applications and are incompatible with essential KV cache optimization techniques like GQA. KV cache quantization \cite{hooper2024kvquant, liu2024kivi, badri2023half}, although practical, does have the upper limit of memory saving by $4-8 \times$.

\textbf{LLM Inference Systems with Offloading.} In modern GPU systems, there is a significant disparity between CPU and GPU memory capacities and costs. CPU RAM typically offers much larger capacity at a lower cost than GPU memory. For example, modern server-grade systems can easily accommodate 1-2TB of CPU RAM, while even high-end GPU cards like the NVIDIA A100 are limited to 80GB of memory. The cost difference is also substantial—server-grade DDR4/DDR5 RAM typically costs a fraction per gigabyte compared to specialized GPU memory. 

This observation is supported by several prominent works offloading the KV cache to the CPU memory: FlexGen \cite{sheng2023flexgen} leverages this hardware characteristic by utilizing CPU memory as an extension of GPU memory, allowing for efficient LLM inference even with limited GPU resources.
DeepSpeed \cite{aminabadi2022deepspeed} implements sophisticated offloading strategies that take advantage of the larger CPU memory capacity to handle model weights and KV cache that would not fit in GPU memory alone. Infinitigen \cite{lee2024infinigen} builds on these foundations by introducing dynamic KV cache management that works synergistically with offloading systems, but its efficiency is highly related to token eviction.

However, this memory hierarchy presents a trade-off: while CPU memory provides larger capacity at lower cost, data transfer between CPU and GPU over PCIe becomes a potential bottleneck due to limited bandwidth compared to GPU memory access speeds. This necessitates careful data movement management and strategic offloading decisions to maintain efficient inference performance.

\subsection{Challenges in KV Cache Management}

In this context, several recent works propose reducing the KV cache size through retrieval head evictions. However, all the prior works assume the persistence of attention patterns across layers, that is, if a head is deemed a retrieval head.

\textbf{KV token eviction affects long-context performance.} Figure \ref{fig:input} shows significant performance degradation since the actual information required by the query might be discarded if considered unimportant, which uses the KV cache of all prior tokens for computing attention results, and the KV cache management method of H2O with a KV cache budget of 2000 tokens. H2O \cite{zhang2024h2o} is a state-of-the-art technique that retains only a small percentage of important tokens in the KV cache to reduce its size. It assesses the importance of each token in every iteration and removes unimportant ones before the next iteration to keep the KV cache size in check.

The figure indicates that this is not the practice case, despite H2O-like approaches assuming that the attention pattern does not change across heads. The tokens deemed unimportant in the one-head iteration could become important in other heads. Consequently, H2O exhibits high similarity until around 200 iterations (i.e., within the KV cache budget). However, as the sequence length extends beyond the KV cache budget, it struggles with the attention pattern's dynamic nature, resulting in more error generation than the optimal case. Note that while we only show the scenario of a KV cache budget of 2000 out of a total sequence length of 10000 tokens for brevity, this issue would become more pronounced as the sequence length surpasses it.

Prior works aiming to reduce the KV cache size through token eviction inherently have some challenges. Given the dynamic attention pattern across iterations, permanently excluding evicted tokens from retinal head token generation can result in a non-negligible drop in accuracy. 
Instead, we must keep the full attention tokens from the retrieval head while selectively evicting less important heads. Furthermore, prior works' iterative allocation of KV cache memory leads to inefficient KV cache management. The number of key/value tokens required increases during chunked-prefill, and each extended context inference demands effective memory management. Failing to account for this allocation may result in ineffective KV cache management. Thus, we need to adjust the memory of key/value token pre-allocation while considering the variances between retrieval and full head.

\section{Roofline Model for head-wise flash attention}

\label{Rooflineappendix}

The Roofline model \cite{williams2009roofline} serves as an effective theoretical framework to assess the potential performance of deploying a model on particular hardware. Here we evaluate hardware performance of memory access and processing unit capabilities. 

\begin{table}[htbp]
\centering
\caption{Performance comparison of different attention mechanisms under RTX-4090 setting}
\small
\begin{tabular}{l|ccccc|cccc}
\hline
\multirow{2}{*}{Operator} & \multicolumn{5}{c|}{Regular} & \multicolumn{4}{c}{Offload} \\
\cline{2-10}
& Ops & Memory & \begin{tabular}[c]{@{}c@{}}Arithmetic\\Intensity\end{tabular} & FLOPS & Bound & \begin{tabular}[c]{@{}c@{}}Memory\\(KV cache)\end{tabular} & \begin{tabular}[c]{@{}c@{}}Arithmetic\\Intensity\end{tabular} & FLOPS & Bound\\
\hline
\multicolumn{9}{l}{\textbf{Prefill}} \\
\hline
flashattention (1k) & 17G & 21M & 820 & 165T & compute & 4.2M & 4100 & 102T & memory \\
flashattention (10k) & 1.7T & 209M & 8200 & 165T & compute & 42M & 41000 & 165T& compute \\
flashattention (100k) & 172T & 2.1G & 82000 & 165T & compute & 419M & 410000 & 165T & compute \\
head-wise  (1k) & 2.1G & 2.6M & 820 & 165T & compute & 0.5M & 4100 & 102T & memory  \\
head-wise  (10k) & 215G & 26M & 8200 & 165T & compute & 5.2M & 41000 & 312T & compute \\
head-wise  (100k) & 21T & 262M & 82000 & 165T & compute & 52M & 410000 & 312T & compute \\
\hline
\multicolumn{9}{l}{\textbf{Decode}} \\
\hline
flashattention (1k) & 17M & 17M & 1 & 1T & memory & 17M & 1 & 13G& memory  \\
flashattention (10k) & 168M & 168M & 1 & 1T & memory & 168M & 1 & 13G& memory  \\
flashattention (100k) & 1.7G & 1.7G & 1 & 1T & memory & 1.7G & 1 & 13G& memory  \\
head-wise  (1k) & 2.1M & 2.1M & 1 & 1T & memory & 2.1M & 1 & 13G& memory  \\
head-wise   (10k) & 21M & 21M & 1 & 1T & memory & 21M & 1 & 13G& memory  \\
head-wise   (100k) & 210M & 210M & 1 & 1T & memory & 210M & 1 & 13G& memory  \\
\hline
\end{tabular}
\label{tab:attention-comparison}
\end{table}

Table \ref{tab:attention-comparison} presents the analysis of layers in Llama-3-8b. From the table, we observe that during the prefill stage, the majority of computations are compute-bound, leading to high performance. Conversely, in the decode stage, all computations are memory-bound, resulting in performance significantly below the computational capacity of the GPU’s computation units. Moreover, offload would make small context prefill memory-bound. Head-wise, the roofline model performs the same arithmetic intensity and peak performance as standard FlashAttention.

\begin{figure*}[h]
\begin{center}
  \includegraphics[scale=0.65]
  {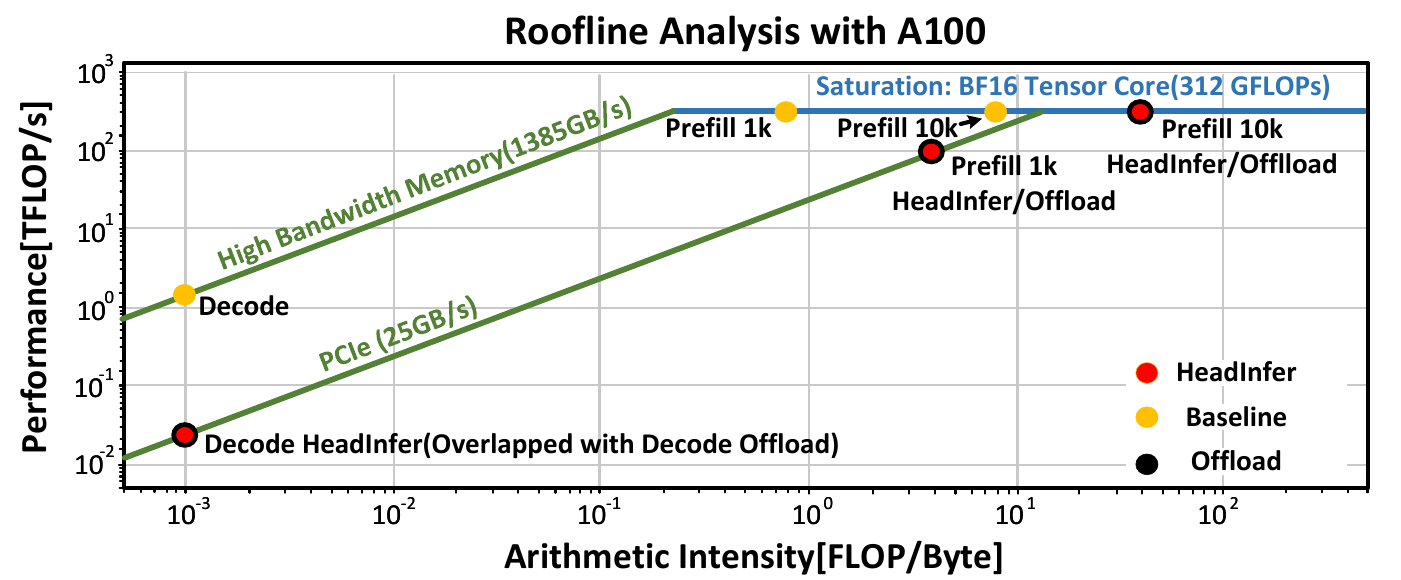}
  \vspace{-1em}
  \caption{Flashattention in the roofline plot analysis using A100 device setting.  }
  \label{fig:roofline_A100}
  \vspace{-1em}
\end{center}
\end{figure*}

We also show the roofline analysis on other GPUs, such as the A100, to demonstrate the generality of this analysis. Figure \ref{fig:roofline_A100} shows that the prefill 1K offload is also positioned on memory-bound while the prefill 10K offload is on the compute-bound. The details data is listed on table \ref{tab:attention-comparison_A100}.

\begin{table}[htbp]
\centering
\caption{Performance comparison of different attention mechanisms under A100 setting}
\small
\begin{tabular}{l|ccccc|cccc}
\hline
\multirow{2}{*}{Operator} & \multicolumn{5}{c|}{Regular} & \multicolumn{4}{c}{Offload} \\
\cline{2-10}
& Ops & Memory & \begin{tabular}[c]{@{}c@{}}Arithmetic\\Intensity\end{tabular} & FLOPS & Bound & \begin{tabular}[c]{@{}c@{}}Memory\\(KV cache)\end{tabular} & \begin{tabular}[c]{@{}c@{}}Arithmetic\\Intensity\end{tabular} & FLOPS & Bound\\
\hline
\multicolumn{9}{l}{\textbf{Prefill}} \\
\hline
flashattention (1k) & 17G & 21M & 820 & 312T & compute & 4.2M & 4100 & 102T & memory \\
flashattention (10k) & 1.7T & 209M & 8200 & 312T & compute & 42M & 41000 & 312T& compute \\
flashattention (100k) & 172T & 2.1G & 82000 & 312T & compute & 419M & 410000 & 312T & compute \\
head-wise  (1k) & 2.1G & 2.6M & 820 & 312T & compute & 0.5M & 4100 & 102T & memory  \\
head-wise  (10k) & 215G & 26M & 8200 & 312T & compute & 5.2M & 41000 & 312T & compute \\
head-wise  (100k) & 21T & 262M & 82000 & 312T & compute & 52M & 410000 & 312T & compute \\
\hline
\multicolumn{9}{l}{\textbf{Decode}} \\
\hline
flashattention (1k) & 17M & 17M & 1 & 1.4T & memory & 17M & 1 & 23G& memory  \\
flashattention (10k) & 168M & 168M & 1 & 1.4T & memory & 168M & 1 & 23G& memory  \\
flashattention (100k) & 1.7G & 1.7G & 1 & 1.4T & memory & 1.7G & 1 & 23G& memory  \\
head-wise  (1k) & 2.1M & 2.1M & 1 & 1.4T & memory & 2.1M & 1 & 23G& memory  \\
head-wise   (10k) & 21M & 21M & 1 & 1.4T & memory & 21M & 1 & 23G& memory  \\
head-wise   (100k) & 210M & 210M & 1 & 1.4T & memory & 210M & 1 & 23G& memory  \\
\hline
\end{tabular}
\label{tab:attention-comparison_A100}
\end{table}

\section{Extension: \sysname Implementation with Head-wise Sparsity}
\label{appx-implementation}
\begin{figure}[t]
\begin{center}
  \includegraphics[scale=0.6]
  {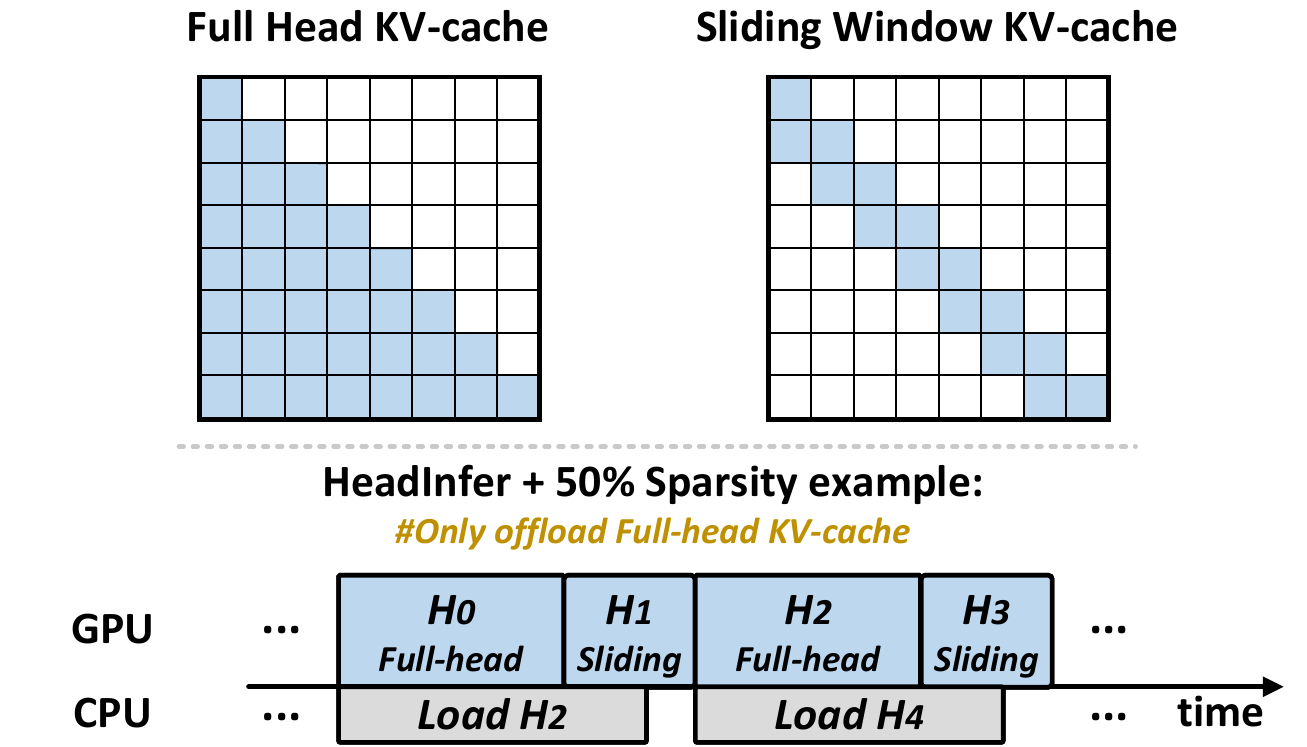}
  \caption{Demonstrations of KV cache policies in inference from the head-wise view. Upper plots illustrate symbolic plots of an attention map deploying different policies in LLM generation. Lower: the overview of \sysname.}
  \label{fig:head-wise sparsity}
  \vspace{-1em}
\end{center}
\end{figure}

Figure \ref{fig:head-wise sparsity} shows our memory management framework. \sysname, which enables offloading the head-wise KV cache with head-wise sparsity. The key design principle behind \sysname is to exploit the redundancy of CPU memory capacity to increase the context size after identifying the important heads in the KV cache. As such, most of the heads for the KV cache are kept in the CPU memory as we generate new tokens, not discarding them like previous work. However, we do not bring the entire KV cache to the GPU for attention but load and compute only the retrieval head of keys and values, leaving other non-retrieval ones staying on the GPU without offloading. To do so, we maintain the head-wise cache pool in the CPU memory and iteratively load the necessary data.

In detail, we use the pre-trained attention input head to speculate the important retrieval head. The speculation is done by processing a customized dataset and analyzing the output. This reduces the waste of PCIe bandwidth by only transferring retrieval heads critical for attention computation. In addition, although the data is offloaded to CPU memory, which is much cheaper and larger than GPU memory, we manage the KV cache pool size so as not to fully utilize the CPU memory.

\subsection{Design Principles}

\textbf{Interleaved KV cache Updates Across GPU and CPU.} The uneven memory consumption and low PCIe link utilization (studied in section 3) during different attention head generation provide an opportunity to exploit the idle GPU memory and PCIe link during the KV cache update phase. To exploit this opportunity, during the attention processing, a head of the attention KV cache can be dynamically fetched on the GPU to compute the attention weight output in parallel while the CPU prefetches the next head. A key requirement to generate the attention output for a given attention head is to stage its parameters (p), query (q), key (k), and value (v), and the attention weight generation is scheduled for each head. In case the key (k) and value (v) of the head are not present on the GPU, the generation operation will trigger a prefetch read from the CPU memory where the head is offloaded, causing I/O operations in the critical execution path of updates. By leveraging the fact that multiple head attention, such as MHA \cite{vaswani2017attention} and GQA \cite{ainslie2023gqa}, are embarrassingly parallel, and \sysname partitions the attention into smaller subgroups, we can perform fine-grained attention generation scheduling across both GPU and CPU without impacting the consistency of generation or introducing computational dependencies between different subgroups. Furthermore, interleaving does not incur memory allocation and deallocation overheads because on the GPU, memory allocation is handled by PyTorch through lightweight memory pools, and on the host, the memory for all subgroups (except static GPU subgroups) is already pre-allocated and pre-pinned (if enabled) during initialization.

\begin{figure*}[t]
\begin{center}
\includegraphics[scale=0.45] 
 {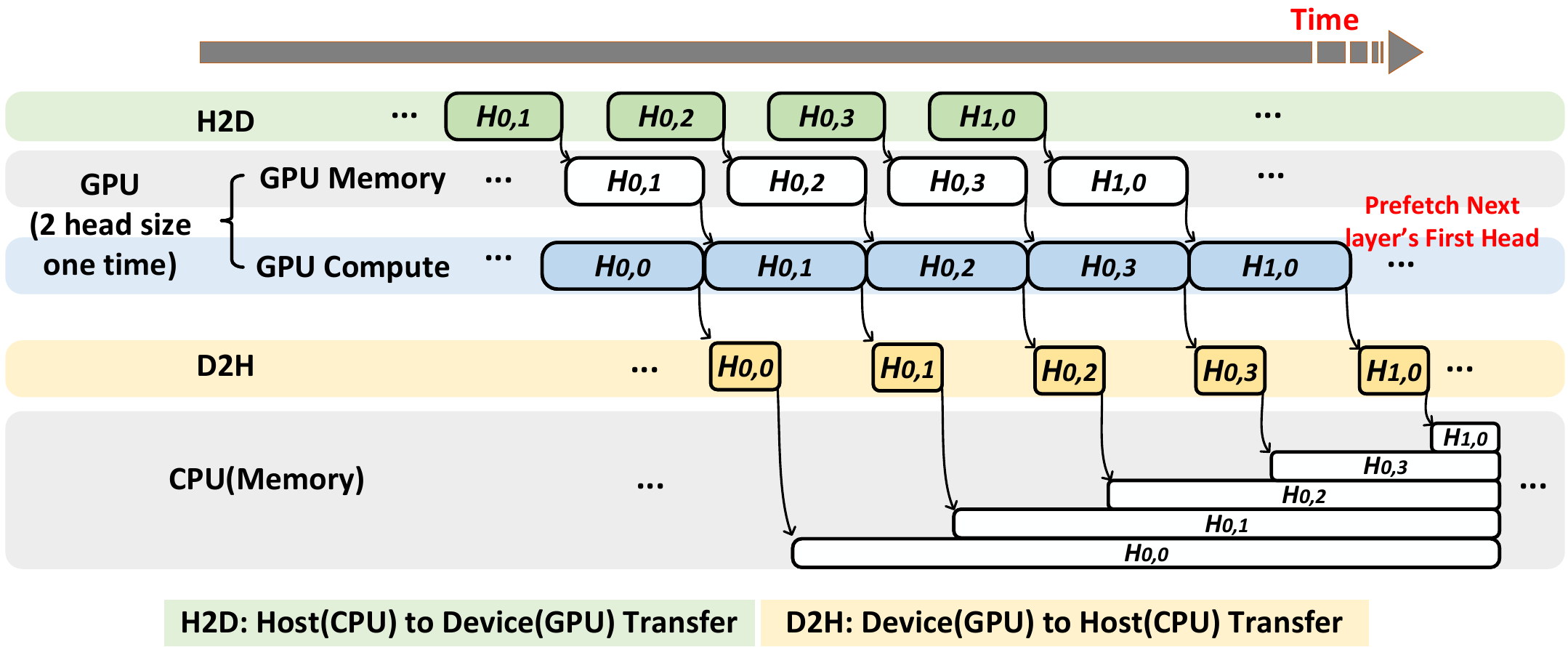}
  \caption{Workflow of \sysname generating a model with (n+1) layers and (j+1) attention heads.}
\label{fig:workflow}
\vspace{-1em}
\end{center}
\end{figure*}
As illustrated in Figure \ref{fig:workflow}, the attention is partitioned into 4 sub-heads, out of which the first head is statically placed on the GPU for the entire inference lifetime; the KV cache corresponding to the head resides in the GPU memory. Therefore, the interleaved offloading adopted by \sysname scheduled all the heads to be updated on the GPU, i.e., for every head updated on the GPU, the host-to-device transform for the next head and the device-to-host transform for the previous head would occur in a non-blocking fashion. This interleaved offloading makes sure that only two heads are maintained on the GPU while most KV cache are offloaded on the CPU. This interleave-centric design allows for efficient overlap between GPU computations and asynchronous head-wise KV cache movement across the PCIe link, which we will detail next.

\textbf{Overlapping Head-wise KV cache Movement and GPU Compute} The data movement observed when the state-of-the-art framework enabling layer-wise KV cache offload (e.g., Flexgen\cite{sheng2023flexgen} and Deepspeed\cite{aminabadi2022deepspeed}) runs an inference process. We observe that after the updates to the KV cache generation to a given layer $i$ are computed on the GPU, the updated KV cache is H2D transferred to the CPU to continue inference in the subsequent chunked-prefill and decoding. Only when all the KV cache of next later are transferred to the GPU can the subsequent iteration begin. Given the parallel nature of attention, the sub-heads can be updated and transferred out of order and do not impact the accuracy of the inference. On contract, using the existing offloading solutions can be slow with head-wise sparsity attributed to (a) the KV cache of all heads within one layer staying on the GPU, which would occupy large memory for long context, and (b) not all attention layers are discretized uniformly, which causes blocking H2D transfer of KV cache.

To mitigate the aforementioned challenges, we propose a head-centric design illustrated in Figure \ref{fig:workflow} for efficient offloading interleaving of H2D transfor and GPU compute. It works as follows: while the GPU computes the generation of the initial head ($H^1$), the KV cache corresponding to the next head ($H^2$), including key (k) and value (v), are being prefetched using asynchronous H2D transfers, thereby overlapping GPU computations with next-head prefetching. Meanwhile, the CPU update for previous $H^0$ is being uploaded using asynchronous D2H transfers. After this, three operations happen in parallel: (1) H2D transfer of the next head and prefetching of the next KV cache to be updated on the GPU (2) Updating of the previous head with KV cache from GPU; and (3) GPU generation of current head outputs, thereby exploiting full-duplex D2H and H2D transfers and parallel GPU computations. 

The attention phase is executed in a parallel fashion by multiple heads of the mechisan. Consequently, our proposed overlapping of GPU computations and PCIe transfers does not incur any blocking overheads, as the computation speed on the GPU is slower than the PCIe throughput to transfer subheads back and forth between the GPU and CPU using high-throughput PCIe.

\textbf{Efficient Management of KV cache}
State-of-the-art hybrid KV cache
offloading solutions (e.g., FlexFLow) by default retain the layer-wise KV cache corresponding to the statically GPU-resident head (h1 to h4) on the GPU during the attention computation, and for the remainder of the layer, KV cache are offloaded to the host memory. We extend this design with head-wise KV cache management to incorporate head-wise sparsity. Head-wise sparsity divides attention heads into two categories: important retrieval heads retain all tokens, while unimportant The streaming head only retains the most recent token. Since the memory usage of the retrieval head will be much larger than that of the streaming head, especially under long context inference, \sysname will selectively offload the retrieval head to the CPU and keep the streaming head in the GPU.

\begin{figure*}[t]
\begin{center}
  \includegraphics[scale=0.5]
  {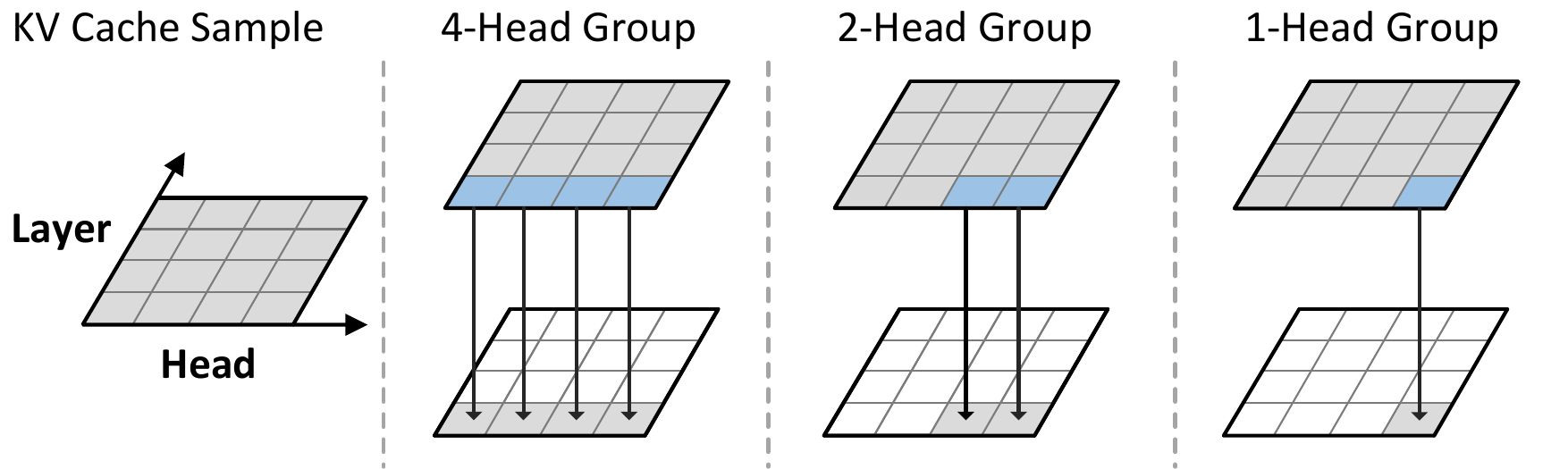}
  \vspace{-0.5em}
  \caption{Demonstrations of adaptive head-wise offloading.}
  \label{fig:adaptive}
  \vspace{-2em}
\end{center}
\end{figure*}

\textbf{Adaptive selecting head-wise granularity}. In transformer-based architectures, attention heads often operate in parallel. However, processing these heads individually—or in small groups—can incur repeated kernel launches and excessive PCIe transfers, particularly if the context size for each head is small. This overhead can quickly dominate total inference time, undermining the benefits of parallelism.
Adaptive head-wise offloading addresses these inefficiencies by merging multiple heads into a single “HeadGroup.” As shown on Figure \ref{fig:adaptive}, with a reduced number of HeadGroups (e.g., 2-HeadGroup or 4-HeadGroup), fewer offloading operations are required, lowering both the latency and the bandwidth usage on the CPU–GPU boundary. Conceptually, this “batching” of heads takes advantage of the fact that many computations within attention heads are structurally similar, thus allowing shared memory transfers and kernel calls.
Layer-offload can be viewed as the extreme case of adaptive head-wise offloading. Instead of grouping only a fraction of heads at a time, layer-offload groups every head in a given layer into one offload operation. While it can drastically reduce overhead further, the trade-off is less granularity and potentially higher intermediate memory requirements. In practice, the decision to use 2-head, 4-head, or full-layer offloading depends on available hardware resources, batch size, and the typical context length being processed.

By carefully tuning the grouping strategy—ranging from small, flexible head groups to large, layer-wide groups—adaptive head-wise offloading makes it possible to significantly optimize inference time in transformer-based models, particularly in latency-sensitive scenarios involving small context.

\subsection{Extension Experiment with Head-wise Sparsity}

We evaluate our head-wise extension of \sysname using 50\% sparsity, which means half of the attention heads are sparse, on Llama3-8B models and compare against \sysname. For the prefill and decoding latency, the extension achieves close to $2\times$ speedup. The results of this evaluation are shown in Table \ref{tab:extension}.

\begin{table}[h]
\centering
\caption{Prefill 1M, Decoding with 1M KV cache performance comparison}
\begin{tabular}{lrr}
\hline
Latency & Prefill 1M (s) & Decoding with 1M KV cache(s)\\
\hline
HeadInfer & 2054 & 6.51 \\
HeadInfer+duoattention 50\% sparsity & 1152 & 3.28 \\
\hline
\end{tabular}
\label{tab:extension}
\end{table}

\section{Easy-to-use and Portable Implementation}
\label{sec:implementation}

\sysname is designed for straightforward integration with existing inference or training frameworks, such as Hugging Face Transformers, requiring only minimal modifications. Below, we illustrate how one can adapt a standard Llama attention module to enable \emph{head-wise} KV offloading with minimal impact on the rest of the code.

\subsection{Overview of Required Modifications} Our changes largely center around intercepting or replacing the attention’s forward pass so that: \begin{itemize}[leftmargin=*] \item \emph{Heads are processed individually or in groups}, rather than as a single large multi-head block. \item \emph{Key and value states} can be transparently stored in, or fetched from, CPU memory (or other off-device storage) before each head’s attention is computed. \item \emph{Asynchronous transfers} are used where possible to overlap CPU--GPU data movement with on-GPU computation. \end{itemize} 

\subsection{Annotated Code Snippet} In the listing below, we demonstrate key functions that illustrate how to integrate \sysname into a \texttt{transformers}-style codebase. These snippets show how the standard \lstinline|forward| method is patched with head-wise logic. We also provide \lstinline|HeadwiseOffloadedCache|, a class that manages CPU--GPU memory movement, and a small helper function for simulating or preparing decoding with large context windows.

The implementation primarily requires modifying the attention mechanism class in transformer models. The key modifications are:

\lstdefinestyle{mypython}{
    language=Python,
    basicstyle=\small,
    keywordstyle=\bfseries\color{blue},
    commentstyle=\itshape\color{teal},
    stringstyle=\color{orange},
    numberstyle=\tiny,             
    numbersep=5pt,                 
    frame=lines,                   
    showstringspaces=false,
    tabsize=4,
    breaklines=true
}

{
\begin{enumerate}
    \item \textbf{Attention Class Modification:} Update the forward pass of the attention mechanism to support head-wise offloading:
    
    \begin{lstlisting}[style=mypython, frame=lines, caption=Modified LlamaAttention forward pass]
def forward(self, hidden_states, attention_mask=None, 
           past_key_value=None):
           
    # Original attention computation
    query_states = self.q_proj(hidden_states)
    key_states = self.k_proj(hidden_states)
    value_states = self.v_proj(hidden_states)
    
    # Head-wise processing
    batch_size = query_states.shape[0]
    num_heads = self.num_heads
    head_dim = self.head_dim
    
    # Reshape for head-wise processing
    query_states = query_states.view(
        batch_size, -1, num_heads, head_dim)
    key_states = key_states.view(
        batch_size, -1, num_heads, head_dim)
    value_states = value_states.view(
        batch_size, -1, num_heads, head_dim)
    
    # Process each head independently
    outputs = []
    for head_idx in range(num_heads):
        head_q = query_states[..., head_idx, :]
        head_k = key_states[..., head_idx, :]
        head_v = value_states[..., head_idx, :]

        # Update or fetch from specialized Cache (past_key_value)
        head_k, head_v = past_key_value.update(head_k, head_v, layer_idx, head_idx)
        
        head_output = compute_head_attention(
            head_q, head_k, head_v, 
            attention_mask,
            past_key_value
        )
        outputs.append(head_output)
    
    return torch.cat(outputs, dim=-1)
    \end{lstlisting}

    \item \textbf{KV Cache Management:} Implement the OffloadedCache class for head-wise cache management:
    
    \begin{lstlisting}[style=mypython, caption=HeadwiseOffloadedCache implementation]
class HeadwiseOffloadedCache:
    def __init__(self, head_groups=1):
        self.key_cache = []
        self.value_cache = []
        self.head_groups = head_groups
        self.prefetch_stream = torch.cuda.Stream()
        self.evict_stream = torch.cuda.Stream()
    
    def update(self, key_states, value_states, layer_idx, head_idx):
        """Updates cache for specific head, managing CPU offload"""
        with torch.cuda.stream(self.prefetch_stream):
            # Prefetch next head if needed
            self._prefetch_head(layer_idx, head_idx + 1)
            
            # Update current head cache
            if len(self.key_cache) <= layer_idx:
                self.key_cache.append([])
                self.value_cache.append([])
                
            while len(self.key_cache[layer_idx]) <= head_idx:
                self.key_cache[layer_idx].append(None)
                self.value_cache[layer_idx].append(None)
                
            # Store on appropriate device
            device = self._get_head_device(layer_idx, head_idx)
            self.key_cache[layer_idx][head_idx] = key_states.to(device)
            self.value_cache[layer_idx][head_idx] = value_states.to(device)
            
        # Evict previous head if needed
        self._evict_head(layer_idx, head_idx - 1)
        
        return key_states, value_states
        
    def _prefetch_head(self, layer_idx, head_idx):
        """Asynchronously moves next head to GPU if needed"""
        if head_idx >= self.num_heads:
            return
        
        cache_k = self.key_cache[layer_idx][head_idx] 
        cache_v = self.value_cache[layer_idx][head_idx]
        
        if cache_k is not None and cache_k.device != torch.device('cuda'):
            cache_k = cache_k.cuda(non_blocking=True)
            cache_v = cache_v.cuda(non_blocking=True)
            
    def _evict_head(self, layer_idx, head_idx):
        """Moves previous head to CPU if needed"""
        if head_idx < 0:
            return
            
        with torch.cuda.stream(self.evict_stream):
            cache_k = self.key_cache[layer_idx][head_idx]
            cache_v = self.value_cache[layer_idx][head_idx]
            
            if cache_k is not None and cache_k.device == torch.device('cuda'):
                cache_k = cache_k.cpu()
                cache_v = cache_v.cpu()
    \end{lstlisting}
\end{enumerate}
}

Our implementation shows that these modifications can be made with minimal changes to existing model architectures while maintaining full compatibility with standard transformer frameworks huggingface. The complete implementation would be available at \url{https://github.com/wdlctc/headinfer}.

\end{document}